\documentclass[final,5p,times,twocolumn]{elsarticle}

\usepackage{adjustbox}
\usepackage{amsmath,amssymb,url}
\usepackage{enumitem}
\usepackage[superscript,,biblabel,noadjust]{cite}
\usepackage{hyperref}
\hypersetup{
    colorlinks=true,
    citecolor=blue,
    linkcolor=red,
    filecolor=magenta,      
    urlcolor=cyan,
}

\usepackage{adjustbox}
\usepackage{multirow}

\setlist[itemize]{leftmargin=*}

\journal{XX}
\begin{document}

\begin{frontmatter}

\title{Modern Views of Machine Learning for Precision Psychiatry}

\author[1,2,3,4]{Zhe Sage Chen\corref{cor1}}
\author[5]{Prathamesh (Param) Kulkarni}%\fnref{fn1}
\author[6,1]{Isaac R. Galatzer-Levy}
\author[1]{Benedetta Bigio}
\author[1,3]{\\Carla Nasca}
\author[7,8]{Yu Zhang\corref{cor1}}
\cortext[cor1]{Corresponding authors: \\
zhe.chen@nyulangone.org (Z.S. Chen, ORCID: 0000-0002-6483-6056)\\
yuzi20@lehigh.edu (Y. Zhang, ORCID: 0000-0003-4087-6544)}
\address[1]{Department of Psychiatry, New York University Grossman School of Medicine, New York, NY 10016, USA}
\address[2]{Department of Neuroscience and Physiology, New York University Grossman School of Medicine, New York, NY 10016, USA}
\address[3]{The Neuroscience Institute, New York University Grossman School of Medicine, New York, NY 10016, USA}
\address[4]{Department of Biomedical Engineering, New York University Tandon School of Engineering, Brooklyn, NY 11201, USA}
\address[5]{Headspace Health, San Francisco, CA 94102, USA}
\address[6]{Meta Reality Lab, New York, NY, USA}
\address[7]{Department of Bioengineering, Lehigh University, PA 18015, USA}
\address[8]{Department of Electrical and Computer Engineering, Lehigh University, PA 18015, USA}

\begin{abstract}

{\bf The bigger picture:} Machine learning (ML) and artificial intelligence (AI) have become increasingly popular in analyzing complex patterns of neural and behavioral data for medicine and psychiatry. We provide a comprehensive review of ML methodologies and applications in precision psychiatry. We argue that advances in ML-powered modern technologies have revolutionalized the current practice in diagnosis, prognosis, monitoring and treatment of various mental illnesses. We discuss conceptual and practical challenges in precision psychiatry and highlight future research in ML.\\

{\bf Summary:} In light of the NIMH’s Research Domain Criteria (RDoC), the advent of functional neuroimaging, novel technologies and methods provide new opportunities to develop precise and personalized prognosis and diagnosis of mental disorders. Machine learning (ML) and artificial intelligence (AI) technologies are playing an increasingly critical role in the new era of precision psychiatry. Combining ML/AI with neuromodulation technologies can potentially provide explainable solutions in clinical practice and effective therapeutic treatment. Advanced wearable and mobile technologies also call for the new role of ML/AI for digital phenotyping in mobile mental health. In this review, we provide a comprehensive review of the ML methodologies and applications by combining neuroimaging, neuromodulation, and advanced mobile technologies in psychiatry practice. Additionally, we review the role of ML in molecular phenotyping, cross-species biomarker identification in precision psychiatry. We further discuss explainable AI (XAI) and causality testing in a closed-human-in-the-loop manner, and highlight the ML potential in multimedia information extraction and multimodal data fusion. Finally, we discuss conceptual and practical challenges in precision psychiatry and highlight ML opportunities in future research.
\end{abstract}


\end{frontmatter}

\section{Introduction}
\label{sec:introduction}
Mental health is epidemic in the United States and the world. According to the National Institute of Mental Health (NIMH), nearly one in five American adults suffer from a form of mental illness or psychiatric disorder (\url{www.nimh.nih.gov/health/statistics/}). According to the Centers for Disease Control and Prevention (CDC), The COVID-19 pandemic has witnessed a significant impact on our lifestyle, and considerably elevated adverse mental health conditions caused by fear, worry and uncertainty \citep{czeisler2020mental}. Increased suicide rates, opioid abuse, antidepressant usage have been observed in both adults and teenagers. The diagnosis and treatment of mental health has imposed a burden to the healthcare system and the society. For instance, the economic burden of depression alone is estimated to be at least \$210 billion annually (\url{www.workplacementalhealth.org/mental-health-topics/depression/}). Precision medicine (or personalized medicine) is an innovative approach to tailoring disease prevention, diagnosis, and treatment that account for the differences in subjects' genes, environments, and lifestyles. The goal of precision medicine is to target timely and accurate diagnosis/prognosis/therapeutics for the individualized patient’s health problem, and further provide feedback information to patients and surrogate decision-makers. Recent decades have witnessed various degrees of successes in precision medicine, especially in oncology (Lancet, vol. 397, 2021, p. 1781). Traditional diagnoses of mental illnesses rely on physical exams, lab tests, psychological and behavioral evaluation. Meanwhile, precision psychiatry has increasingly received its deserved attention \citep{insel2015brain,fernandes2017new}. Although psychiatry has not yet benefited fully from the advanced diagnostic and therapeutic technologies that have an impact on other clinical specialties, these technologies have the potential to transform the future psychiatric landscape.

The NIMH’s RDoC (Research Domain Criteria) initiative aims to address the heterogeneity of mental illness and provide a biology-based (as opposed to symptom-based) framework for understanding these mental illnesses in terms of varying degrees of dysfunction in psychological or neurobiological systems; it attempts to bridge the power of multidisciplinary (such as the genetics, neuroscience, and behavioral science) research approaches \citep{insel2010research}. The current gold standard for diagnosis and treatment outcome in mental disorders---the DSM (Diagnostic and Statistical Manual of Mental Disorders) maintained by the American Psychiatric Association (APA), are often based on the clinician’s observations, behavioral symptoms and patient reporting, which are all susceptible to a high degree of variability. Therefore, it is imperative to develop quantitative neurobiological markers for mental disorders while accounting for their heterogeneity and comorbidity.

\begin{figure}[!t]
\centering
\includegraphics[width=9cm]{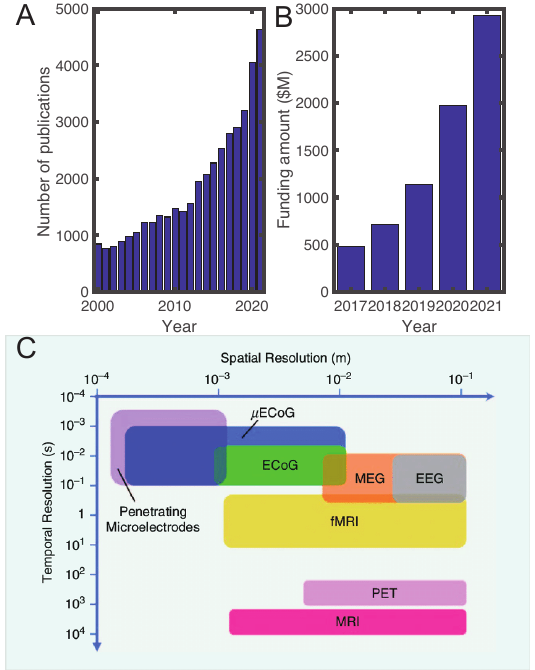}
\caption{(A) The number of PubMed publications with keywords ''machine Learning or AI'' and ''psychiatry or mental health'' in the title or abstract (Year 2000-2021).
(B) Growth of mental health tech funding in the US market  (Year 2017-2021, data source: cbinsights.com). 
(C) Human neuroimaging at various spatial and temporal resolution ($\copyright$ IEEE, figure reproduced from \citep{thukral2018IEEE} with permission). }
\label{fig1}
\end{figure}

One important goal in neuropsychiatry research is to identify the relationship between neurobiological/neurophysiological findings and clinical behavioral/self-report observations. Machine learning (ML) and artificial intelligence (AI) have generated growing interests in psychiatry because of their strong predictive power and generalization ability for prognosis and diagnosis applications \citep{bzdok2018machine,zhou2020machine,allen2020synthesising}. The interest of applying ML/AI in psychiatry has grown steadily in the past two decades, as reflected in the number of PubMed publications (Figure~\ref{fig1}A). To improve mental health outcome with digital technologies, the so-called ``digital psychiatry" focuses on developing ML/AI methods for assessing, diagnosing, and treating mental health issues \citep{burr2020digital}. A recent global survey has indicated that psychiatrists were somewhat skeptical that AI could replace human empathy, but many predicted that 'man and machine' would increasingly collaborate in undertaking clinical decisions, and psychiatrists  were optimistic that AI might improve efficiencies and access to mental care, and reduce costs \citep{doraiswamy2020artificial}.

The past two decades have witnessed substantial growth of ML applications for psychiatry in the literature, reflected in many applications and reviews \citep{rutledge2019machine,chandler2020using,galatzer2018data,shatte2019machine,su2020deep,liu2020brief,thieme2020machine,durstewitz2019deep,koppe2021deep,hedderich2021,bracher2021}. Although multiple reviews of ML for psychiatry are available, the majority of reviews are restricted to relatively narrow scopes. In this paper, we try to provide a comprehensive review of ML and ML-powered technologies in mental health applications. Our view is ``modern'' in a sense that the development of new technologies, consumer market demand, and public health crises (such as COVID-19) have constantly redefined the role of ML and reshaped our thinking in precision psychiatry. Specifically, we will cover state-of-the-art technological and methodological developments in ML, multimodal neuroimaging, neuromodulation, large-scale circuit modeling and human-machine interface. It is noteworthy that our reviewed literature is by no means exhaustive due to space limitation. To distinguish our review from others, we will focus on several issues central to the ML applications for psychiatry: generalizability, interpretability, causality, clinical and behavioral integration.

Our view about this emerging field is optimistic for several reasons: first, with increasing amount of data and computational power, there is a growing  demand for psychiatrists to use ML to reevaluate  clinical, behavioral and neuroimaging data. The interests in mental health funding from the industry have also grown substantially  (Figure~\ref{fig1}B). Second, it is becoming increasingly important to leverage the power of ML and develop explainable artificial intelligence (XAI) tools for unbiased risk diagnosis, personalized medicine recommendation, and neurostimulation. The integration of ML with advanced neuroimaging can potentially help us identify and validate biomarkers in diagnosis and treatment of mental illnesses. Third, there is an increasing demand of psychiatrists in the US, and the shortage is even more acute in poorer countries \citep{allen2020artificial}. ML/AI technologies may change the practice of psychiatry for both clinicians and patients. Finally, advanced technologies such as social media, multimedia (speech and vision), mobile and wearable devices also call for the development of ML/AI tools to assist the assessment, diagnosis or treatment of individuals who are mentally ill or at risk. From now on, we will use ML and AI interchangeably throughout the paper.

\section{Background of Neuroimaging}
\subsection{Mind vs. Brain}
The brain is a physical organ, which provides the center that supports all cognitive functions. It can be considered as the hardware of the human body. In contrast, the mind is abstract; it creates emotions and enables consciousness, perception, thinking, judgment, and memory. The World Health Organization (WHO) defines mental health as ''\textit{a state of well-being in which the individual realizes his or her own abilities, can cope with the normal stresses of life, can work productively and fruitfully, and is able to make a contribution to his or her community.}" Psychiatry is seeking to measure the mind, and relies on the quantification of how people feel under specific tasks or behavioral conditions. Psychiatric disorders are often described as disorders of the mind, which disrupt the brain’s ability to function normally to complete certain processes. For instance, anxiety, stress, depression, autism, obsessive-compulsive disorder (OCD), and post-traumatic stress disorder (PTSD) are categorized by varying degrees of psychomotor, cognitive, affective, and volitional impairment. Since the brain and mind are internally interleaved, the syndromes of mental disorders are associated with dysfunctions of neural, cognitive, and behavioral systems \citep{insel2015brain}. To unravel mysteries of the mind, we need to understand the brain based on modern neuroimaging techniques.

\subsection{Advances in Neuroimaging}
Neuroimaging provides a window to probe human brains in terms of both structural and functional forms, and offers various resolutions to examine brain activity at macroscopic, mesoscopic, and microscopic scales across spatial and temporal domains (Figure~\ref{fig1}C). 

Our understanding of brain and behavior relationships has expanded exponentially over the last few decades. While this improvement may be attributed to a multitude of factors, advancement in neuroimaging has played a prominent role \citep{etkin2019reckoning}. Ranging from increased utilization of structural neuroimaging techniques to the significant scientific advancements brought about by the increased availability of functional neuroimaging, these technologies have provided significant benefits to improved understanding of neural correlates and discovery of biomarkers in psychiatric disorders \citep{noda2020neural,noggle2021advances}. Some of the most common neuroimaging methods for probing brain function include the utilization of magnetic resonance imaging (MRI), functional MRI (fMRI), diffusion tensor imaging (DTI), electroencephalography (EEG), Magnetoencephalography (MEG), electrocorticography (ECoG), functional near-infrared spectroscopy (fNIRS), and positron emission tomography (PET).

\begin{itemize}
\item \textbf{MRI.} MRI is a non-invasive imaging technology that produces 3-D anatomical images and has been widely used in clinics for diagnosis, staging, and follow-up of brain disease \citep{olabi2011there}.
\item \textbf{DTI.} DTI is a technological advancement that allows researchers and clinicians to assess human white matter pathways in vivo \citep{podwalski2021magnetic}. DTI has shown its promise in characterizing neuroanatomical connectivity underlying white matter tracts, which is hard to be captured by MRI.
\item \textbf{fMRI.} fMRI is a hemodynamic technique to characterize functional activity of the brain by measuring blood oxygenation level-dependent (BOLD) signals \citep{smith2013resting}. Due its good spatial resolution, fMRI has been extensively used for studying neurobiological basis of various psychiatric disorders.
\item \textbf{PET.} PET is a functional imaging technique that uses radioactive substances known as radiotracers to visualize and measure changes in metabolic processes, and other physiological activities including blood flow, regional chemical composition, and absorption. The successful application of PET has been reported in biomarker studies for psychotic disorders \citep{coughlin2019opportunities}.
\item \textbf{EEG.} EEG measures electrical activity from neuronal populations via electrodes on the scalp. Due to its low cost and ease of data collection, EEG provides a practical tool to be used in a variety of clinical environments \citep{hamilton2020electroencephalography}.
\item \textbf{MEG.} MEG uses superconducting quantum interference devices to measure the magnetic fields produced by electrical activity in the brain. MEG is less affected by secondary currents and demonstrates superior spatial resolution compared with EEG \citep{williams2018recent}.
\item \textbf{ECoG.} ECoG can be viewed as invasive intracranial EEG that measures brain signals from the cortical surface. Because the subdural electrodes are placed directly on the cortical surface, the ECoG signals have better signal-to-noise ratio (SNR) and excellent spatial and spectral resolution compared with EEG \citep{parvizi2018}. ECoG has a long history of use in clinical neurosurgery since its initial application in epilepsy surgery, and has been recently used in research studies of mental disorders, such as depression and mood disorders \citep{sani2018mood,scangos2021}. Micro-ECoG ($\mu$ECoG) utilizes micro-scale electrodes with contact site diameters many orders of magnitude smaller than traditional clinical ECoG electrode sites and minimized inter-electrode spacing, allowing for even greater spatial resolution of measured brain signals.
\item \textbf{fNIRS.} fNIRS is an optical brain monitoring technique that uses near-infrared spectroscopy to measure cortical hemodynamic activities for functional neuroimaging \citep{ferrari2012brief}. Alongside EEG, fNIRS can also be used in portable settings for studying psychiatric disorders \citep{gosse2021functional}.
\end{itemize}

To date, EEG and fMRI are two most commonly used modalities for precision psychiatry. Specifically, EEG is low-cost and easy-to-operate, making it more appealing for psychiatric practice or home use.

\subsection{Neuroimaging Analysis}
The rich neuroimaging modalities allow us to comprehensively probe brain functions. Numerous research efforts have been devoted to revealing the neurobiological basis of various psychiatric disorders using advanced neuroimaging analyses, which have focused on the following aspects.

\begin{itemize}
\item \textbf{Task-related brain activation.} Under specifically designed cognitive paradigms, the collected neuroimaging data enable us to examine brain activities associated with certain experimental tasks and study their relationship with cognitive dysfunctions. Typical measures of task-related brain activities include event-related potential (ERP) and event-related spectral perturbation, and reward or emotional processing-related functional activation \citep{keren2018reward,lukow2021neural}.
\item \textbf{Brain structural and functional connectivity.} A promising direction for probing brain function using neuroimaging is to investigate brain connectivity (or connectome) \citep{smith2013functional}. Studying the resting-state brain connectome provides a promising way to characterize the complex brain architecture and uncover brain dysfunctions in intrinsic brain networks \citep{woodward2015resting}.
\item \textbf{Brain dynamics.} Increasing neuroimaging studies suggest that functional connectivity may fluctuate, rather than being stationary during an entire session of data collection \citep{ma2014dynamic}. Studies examining spatiotemporal dynamics of brain networks have recently received increasing attention and may reveal meaningful brain states associated with different psychiatric conditions \citep{rolls2021brain}.
\item \textbf{Multimodal neuroimaging.} Another promising approach to establish robust biomarkers for psychiatry is to combine multiple neuroimaging modalities, which offers opportunities to exploit cross-modality complementary information that a single modality approach may not capture \citep{calhoun2016multimodal}.
\end{itemize}

To fully understand the mind, we argue that neuroimaging, when combined with modern ML and other ML-powered technologies, can provide powerful tools in advancing diagnosis, prognosis, and intervention of psychiatric disorders.

\section{What and How ML Can Help Psychiatry?}
\label{section-WhatHow}
\subsection{What Makes Psychiatry Different from Other Medicine Disciplines?}
The nature and etiology of mental illnesses remain unclear and challenging to study. Psychiatric disorders are typically diagnosed according to a combination of clinical symptoms based on the Diagnostic and Statistical Manual of Mental Disorders (DSM) or the International Classiﬁcation of Diseases (ICD). Traditional studies for the neurobiology of psychiatric disorders have followed a categorical classification framework using a case-control design whereby all patients with a given diagnosis are compared with healthy individuals. This framework largely relies on clinically derived diagnostic labels, assuming that a biomarker may differ enough between healthy people and patients. The symptom-based diagnosis covered hundreds of thousands of different symptom combinations, which has caused extensive clinical heterogeneity \citep{feczko2019heterogeneity,satterthwaite2020parsing}. It is increasingly recognized that existing clinical diagnostic categories could misrepresent the causes underlying mental disturbance. The traditional case-control design has limited strengths in delineating the significant clinical and neurobiological heterogeneity of psychiatric disorders, thereby hindering the understanding of psychopathology and the search for biomarkers. On the other hand, previous studies have broadly explored the group effects of neurobiology to explain its connection to behavior and disease. Such group-level analyses cannot fully capture individual-level brain abnormality that is crucial for developing personalized medicine. In addition, many psychiatric disorders may be considered as falling along multiple dimensions. Co-occurrence of multiple psychiatric disorders might reflect different patterns of symptoms resulting from shared risk factors and perhaps the same underlying disease processes. The high comorbidity in these disorders significantly affects the characterization of psychopathology according to the traditional diagnostic categories. Conventional studies focusing on a single diagnostic domain are therefore limited in uncovering the neural correlates of comorbidity among multiple disorders and identifying the dimensions of neural circuits and behavioral phenotypes.

\begin{table*}
\centering
    \caption{Categories of ML, concepts, typical methods, and their representative applications.}
  \scalebox{0.8}{
  \begin{tabular}{llll}
  \hline

\bf Learning category &	\bf Concepts  &	\bf Representative methods & \bf Applications \\ \hline
 
 Supervised  & Learning from labeled data  & SVM, random forest, & Disease diagnosis, prognosis, \\
 
 &  to predict class/clinical measures & sparse learning, ensemble learning  &   treatment outcome prediction\\ \hline
 
  Unsupervised & Learning from unlabeled data to uncover  & 
Hierarchical clustering,  & Disease subtyping, normative   \\

& structure and identify subgroups & K-means, PCA, CCA &  modeling, identify behavioral \\
& & & and neurobiological dimension \\ \hline
  
 Semi-supervised  & Learning from both labeled and  & Multi-view learning,
  & Multimodal analysis, Joint \\

& unlabeled data to perform supervised  & Laplacian regularization,  &  disease subtyping and diagnosis,\\
& or unsupervised tasks & Semi-supervised clustering & Prediction with incomplete data \\ \hline

 Deep  & Learning hierarchies and nonlinear mappings  & CNN, Deep autoencoder, GCN, &  A large class of generic  \\
   & of features for higher-level representations  & RNN, LSTM, GAN & learning problems \\
  \hline
\end{tabular}}
\label{tab_ML}
\end{table*}

Distinct from the traditional case-control design, the NIMH’s RDoC aims to address the heterogeneity and comorbidity in psychiatry by linking symptom dimensions with biological systems, cutting across the diagnostic spectrum \citep{insel2014nimh}. The ultimate goal of RDoC is to ﬁnd ``new ways of classifying psychiatric diseases based on multiple dimensions of biology and behavior" \citep{cuthbert2014rdoc}. These newly defined disease dimensions could further be utilized to discover neurobiological phenotypes and clarify the causal mechanism underlying the associated brain dysfunctions. To achieve this important vision, new analytical approaches are urgently needed. Thanks to the advancement in cutting-edge  ML/AI techniques, psychiatrists and investigators can beneﬁt from a deep understanding of complex patterns in brain, behavior, and genes \citep{bzdok2018machine}. Combining these analysis techniques with rich multimodal data from increasing large-scale multi-center cohorts holds significant promise in advancing a biologically grounded redeﬁnition of psychiatric disorders.

Despite rapid progress in psychiatric studies, several areas appear highly underexplored but may carry the substantial potential for achieving major breakthroughs toward precision psychiatry. The capacity to dissect inter- and intra-individual variability is crucial for better understanding the neural basis of variation in human cognition and behavior \citep{wang2015parcellating}. Studies focusing on the level of the individual may find greater success over conventional group-level analyses. Translational study-orientated approaches for psychiatric neuroimaging may further enhance the ability to ﬁnd statistically significant effect sizes that can be used in individuals \citep{walter2019translational}. On the other hand, identifying subgroups (i.e., subtypes) in psychiatric disorders offers a promising way to delineate disease heterogeneity. Increasing evidence suggests that data-driven subtyping may drive novel neurobiological phenotypes associated with distinctive behavior and cognitive functioning \citep{drysdale2017resting}. These stratified phenotypes may further show improved predictability for clinical outcomes than DSM/ICD diagnoses and serve as potential markers for treatment selection \citep{zhang2021identification}. Another promising area focuses on transdiagnostic approaches to uncover neural correlates of specific domains, such as cognition, arousal, and emotion regulation, which are implicated in psychopathology across the diagnostic spectrum \citep{sargent2021resting}. Recent ML efforts have been dedicated to identifying transdiagnostic brain dysfunctions and dimensions of psychopathology \citep{barch2017neural,xia2018linked,kebets2019somatosensory,mcteague2020identification}. Importantly, leveraging ``big data" from a longitudinal perspective offers a promising way to track the neurobiological and phenotypic trajectories that have been rarely examined in previous cross-sectional studies of psychiatric disorders \citep{wachinger2018longitudinal,vidal2019bidirectional,roeckner2021neural}. Such longitudinal studies may help reveal the neural mechanism underlying the disease progression, and provide new insights for the development of timely interventions. These new frontiers in studying psychiatric disorders can be substantially empowered by ML methodologies summarized in Table~\ref{tab_ML}, including stratifying patients into clinically meaningful subtypes, discovering novel transdiagnostic disease dimensions, and tailoring treatment decisions to individual patients. The research outcome can deliver a significant promise in promoting the development of objective biomarkers-based precision psychiatry.

The applications of ML in psychiatry can be mainly categorized as four types: diagnosis, prognosis, treatment, and readmission. In contrast to most medical disciplines, traditional diagnoses in psychiatry remain restricted to subjective symptoms and observable signs, and therefore call for a paradigm shift. In the following subsections, we will review several key ML paradigms in mental health applications based on neuroimaging, behavioral and clinical measurements. A tabular review of representative applications is shown in Table~\ref{tab_ML_app}. In this section, we focus on the review of neuroimaging-based psychiatric studies, and detailed reviews of the other data domains (such as genetic, clinical, behavioral, and social media data) will be presented in later sections.

\begin{table*}
\centering
    \caption{Representative ML applications in psychiatry based on neuroimaging and clinical data.}
 \scalebox{0.65}{
  \begin{tabular}{lllll}
  \hline

\bf Application &	\bf Method &	\bf Disease & \bf Data type & \bf Reference \\ \hline
 
\multirow{16}{*}{Diagnosis} & \multicolumn{1}{l}{Classification (dynamic GCN)} & \multicolumn{1}{l}{ADHD} & \multicolumn{1}{l}{rs-fMRI + Phenotypic data} & \multicolumn{1}{l}{\citep{zhao2022dynamic}}  \\\cline{2-5}

  & \multicolumn{1}{l}{Classification (Ensemble learning)} & \multicolumn{1}{l}{ADHD}      & \multicolumn{1}{l}{Multimodal} & \multicolumn{1}{l}{\citep{luo2020multimodal}} \\\cline{2-5}
  
   & \multicolumn{1}{l}{Classification (GCN)} & \multicolumn{1}{l}{ASD}      & \multicolumn{1}{l}{Task fMRI} & \multicolumn{1}{l}{\citep{li2021braingnn}} \\\cline{2-5}

   & \multicolumn{1}{l}{Classification (Ensemble learning + GCN)} & \multicolumn{1}{l}{ASD}      & \multicolumn{1}{l}{rs-fMRI} & \multicolumn{1}{l}{\citep{khosla2019ensemble}} \\\cline{2-5}
   
    & \multicolumn{1}{l}{Classification (PCA + LASSO)} & \multicolumn{1}{l}{Bipolar}      & \multicolumn{1}{l}{DWI + Cognitive data} & \multicolumn{1}{l}{\citep{wu2017identification}} \\\cline{2-5} 
    
    & \multicolumn{1}{l}{Classification (RVM)} & \multicolumn{1}{l}{PTSD}      & \multicolumn{1}{l}{rs-fMRI} & \multicolumn{1}{l}{\citep{zhu2020multivariate}} \\\cline{2-5}     

    & \multicolumn{1}{l}{Classification (ICA + LSTM)} & \multicolumn{1}{l}{Schizophrenia}      & \multicolumn{1}{l}{fMRI} & \multicolumn{1}{l}{\citep{yan2019discriminating}} \\\cline{2-5}  
    
    & \multicolumn{1}{l}{Classification (SVM)} & \multicolumn{1}{l}{Schizophrenia}      & \multicolumn{1}{l}{sMRI} & \multicolumn{1}{l}{\citep{mikolas2018machine}} \\\cline{2-5}    
    
     & \multicolumn{1}{l}{Classification (CNN)} & \multicolumn{1}{l}{Depression}      & \multicolumn{1}{l}{rs-EEG} & \multicolumn{1}{l}{\citep{uyulan2021major}} \\\cline{2-5}    
      & \multicolumn{1}{l}{Classification (Autoencoder + MLP)} & \multicolumn{1}{l}{ASD}      & \multicolumn{1}{l}{rs-fMRI} & \multicolumn{1}{l}{\citep{heinsfeld2018identification}} \\\cline{2-5}      
     
      & \multicolumn{1}{l}{Classification (GNN)} & \multicolumn{1}{l}{ASD}      & \multicolumn{1}{l}{rs-fMRI + Phenotypic data} & \multicolumn{1}{l}{\citep{parisot2018disease}} \\\cline{2-5}
      
      & \multicolumn{1}{l}{Subtyping (Normative modeling + clustering)} & \multicolumn{1}{l}{PTSD}      & \multicolumn{1}{l}{rs-fMRI} & \multicolumn{1}{l}{\citep{maron2020individual}} \\\cline{2-5} 
      
      & \multicolumn{1}{l}{Subtyping (CCA + Hierarchical clustering)} & \multicolumn{1}{l}{Depression}      & \multicolumn{1}{l}{rs-fMRI} & \multicolumn{1}{l}{\citep{drysdale2017resting}} \\\cline{2-5}
      
      & \multicolumn{1}{l}{Subtyping (Sparse K-means)} & \multicolumn{1}{l}{PTSD and Depression}      & \multicolumn{1}{l}{rs-EEG} & \multicolumn{1}{l}{\citep{zhang2021identification}} \\\cline{2-5} 
      
    & \multicolumn{1}{l}{Subtyping (Latent class analysis)} & \multicolumn{1}{l}{ADHD}      & \multicolumn{1}{l}{Task fMRI} & \multicolumn{1}{l}{\citep{lecei2019can}} \\\cline{2-5} 
 
      & \multicolumn{1}{l}{Transdiagnostic (Normative modeling + GP regression)} & \multicolumn{1}{l}{Multiple disorders}      & \multicolumn{1}{l}{rs-fMRI} & \multicolumn{1}{l}{\citep{parkes2021transdiagnostic}} \\\cline{2-5}
      
     & \multicolumn{1}{l}{Transdiagnostic (Sparse CCA)} & \multicolumn{1}{l}{Multiple disorders}      & \multicolumn{1}{l}{rs-fMRI} & \multicolumn{1}{l}{\citep{xia2018linked}} \\\cline{2-5} 
     
     & \multicolumn{1}{l}{Transdiagnostic (PLS)} & \multicolumn{1}{l}{Multiple disorders}      & \multicolumn{1}{l}{rs-fMRI} & \multicolumn{1}{l}{\citep{kebets2019somatosensory}} \\\cline{2-5} \hline
     
\multirow{8}{*}{Prognosis} & \multicolumn{1}{l}{SVM} & \multicolumn{1}{l}{Psychosis, Depression} & \multicolumn{1}{l}{multimodal} & \multicolumn{1}{l}{\citep{koutsouleris2021multimodal}}  \\\cline{2-5}     
      & \multicolumn{1}{l}{LASSO} & \multicolumn{1}{l}{Psychosis}      & \multicolumn{1}{l}{rs-EEG} & \multicolumn{1}{l}{\citep{ramyead2016prediction}} \\\cline{2-5} 
 
       & \multicolumn{1}{l}{SVM} & \multicolumn{1}{l}{Depression}      & \multicolumn{1}{l}{rs-EEG} & \multicolumn{1}{l}{\citep{zhdanov2020use}} \\\cline{2-5}   
       
       & \multicolumn{1}{l}{GP classifier} & \multicolumn{1}{l}{Depression}      & \multicolumn{1}{l}{Task fMRI} & \multicolumn{1}{l}{\citep{schmaal2015predicting}} \\\cline{2-5}      
       
       & \multicolumn{1}{l}{LASSO} & \multicolumn{1}{l}{Substance use}      & \multicolumn{1}{l}{MRI/task fMRI} & \multicolumn{1}{l}{\citep{bertocci2017reward}} \\\cline{2-5}     
              
        & \multicolumn{1}{l}{LSTM} & \multicolumn{1}{l}{PTSD}      & \multicolumn{1}{l}{MEG} & \multicolumn{1}{l}{\citep{zhang2020predicting}} \\\cline{2-5}           
       
         & \multicolumn{1}{l}{DNN} & \multicolumn{1}{l}{PTSD}      & \multicolumn{1}{l}{rs-fMRI / task fMRI} & \multicolumn{1}{l}{\citep{sheynin2021deep}} \\\cline{2-5}        
 
          & \multicolumn{1}{l}{SVM} & \multicolumn{1}{l}{Schizophrenia}      & \multicolumn{1}{l}{sMRI} & \multicolumn{1}{l}{\citep{nieuwenhuis2017multi}} \\\cline{2-5}   
          
          & \multicolumn{1}{l}{MLP} & \multicolumn{1}{l}{Schizophrenia}      & \multicolumn{1}{l}{Task fMRI} & \multicolumn{1}{l}{\citep{smucny2021comparing}} \\\cline{2-5} \hline 
          
 \multirow{8}{*}{Treatment } & \multicolumn{1}{l}{Latent space learning} & \multicolumn{1}{l}{Depression} & \multicolumn{1}{l}{rs-EEG} & \multicolumn{1}{l}{\citep{wu2020electroencephalographic}}  \\\cline{2-5}  
 
 \multirow{8}{*}{prediction }          & \multicolumn{1}{l}{RVM} & \multicolumn{1}{l}{Depression}      & \multicolumn{1}{l}{Task fMRI} & \multicolumn{1}{l}{\citep{fonzo2019brain}} \\\cline{2-5}

             & \multicolumn{1}{l}{SVM  } & \multicolumn{1}{l}{Psychosis}      & \multicolumn{1}{l}{ sMRI} & \multicolumn{1}{l}{\citep{koutsouleris2009use}} \\\cline{2-5}     
 
            & \multicolumn{1}{l}{SVM + GP classifier } & \multicolumn{1}{l}{Depression}      & \multicolumn{1}{l}{ sMRI} & \multicolumn{1}{l}{\citep{redlich2016prediction}} \\\cline{2-5}            
           
           & \multicolumn{1}{l}{SVM } & \multicolumn{1}{l}{ADHD}      & \multicolumn{1}{l}{ sMRI} & \multicolumn{1}{l}{\citep{chang2021regional}} \\\cline{2-5}           
           
             & \multicolumn{1}{l}{GP classifier } & \multicolumn{1}{l}{PTSD}      & \multicolumn{1}{l}{ MRI/rs-fMRI} & \multicolumn{1}{l}{\citep{zhutovsky2019individual}} \\\cline{2-5}            
             
              & \multicolumn{1}{l}{MVPA regression } & \multicolumn{1}{l}{ASD}      & \multicolumn{1}{l}{ Task fMRI} & \multicolumn{1}{l}{\citep{yang2016brain}} \\\cline{2-5}            
 
            & \multicolumn{1}{l}{LASSO } & \multicolumn{1}{l}{Anxiety}      & \multicolumn{1}{l}{ rs-fMRI} & \multicolumn{1}{l}{\citep{reggente2018multivariate}} \\\cline{2-5} 
            
               & \multicolumn{1}{l}{SVM } & \multicolumn{1}{l}{Schizophrenia}      & \multicolumn{1}{l}{ rs-fMRI} & \multicolumn{1}{l}{\citep{cao2020treatment}} \\\cline{2-5} \hline   
               
  \multirow{4}{*}{Readmission} & \multicolumn{1}{l}{SVM} & \multicolumn{1}{l}{Depression} & \multicolumn{1}{l}{multimodal} & \multicolumn{1}{l}{\citep{cearns2019predicting}}  \\\cline{2-5}                
  
    & \multicolumn{1}{l}{Classification tree } & \multicolumn{1}{l}{Bipolar}      & \multicolumn{1}{l}{ EHR} & \multicolumn{1}{l}{\citep{edgcomb2019high}} \\\cline{2-5}              
    & \multicolumn{1}{l}{Ensemble learning } & \multicolumn{1}{l}{Substance use}      & \multicolumn{1}{l}{ Phenotypic data} & \multicolumn{1}{l}{\citep{morel2020predicting}} \\\cline{2-5}   
    
    & \multicolumn{1}{l}{Growth mixture modeling } & \multicolumn{1}{l}{Depression}      & \multicolumn{1}{l}{ Clinical data} & \multicolumn{1}{l}{\citep{gueorguieva2017trajectories}} \\\cline{2-5}       
  \hline
\end{tabular}}
\label{tab_ML_app}
\end{table*}

\subsection{Supervised and Unsupervised Learning}
ML holds substantial promise in promoting research from small case-control studies to those with large transdiagnostic samples, and from prior specified brain regions to whole-brain circuit dysfunction for individual-level precision medicine \citep{cearns2019recommendations,etkin2019reckoning,grzenda2021evaluating}. In a new era of evidence-based psychiatry tailored to individual patients, objectively measurable endophenotypes could allow for early disease detection, personalized treatment selection, and dosage adjustment to reduce the burden of disease \citep{rutledge2019machine,tai2019machine,aafjes2021scoping}. These promising applications in psychiatric disorders have been enabled by leveraging the powerful strength of modern ML techniques \citep{shatte2019machine,nielsen2020machine,sui2020neuroimaging,zhang2020survey}.

\textbf{Supervised Learning.} Supervised learning, as the most popularly used category, has been widely applied to neuroimaging-based predictive modeling tasks for psychiatric disorders \citep{cho2019review}. Classic supervised learning algorithms include logistic regression, support vector machine, and random forest. Given the high-dimensional nature of neuroimaging data, these approaches are commonly accompanied by a feature selection step to obtain low-dimensionality representations.  Connectome-based predictive modeling \citep{finn2015functional,shen2017using} is one of such approaches that combine simple linear regression and feature selection to predict individual differences in traits and behavior from connectivity data. LASSO provides an alternative approach that performs simultaneous feature selection and prediction to learn a compact feature pattern for the accurate prediction of a specific brain disease or clinical outcome \citep{reggente2018multivariate}. Relevance vector machine (RVM) builds upon a probabilistic framework by leveraging automatic relevance determination to learn a sparse solution and penalize unnecessary complexity in the model \citep{tipping2001sparse,zhang2015sparse}. RVM has recently demonstrated its strength in quantifying neuroimaging biomarkers for PTSD diagnosis \citep{zhu2020multivariate} and for treatment outcome prediction in depression \citep{fonzo2019brain}. As an extension of the conventional single-task methods, multi-task learning approaches have been increasingly employed to exploit complementary features jointly from multiple views of neuroimaging data \citep{ma2018classification,xiao2019manifold,kim2022multi}.

Due to the complex nature of brain function, informative features may not be observable in the raw high-dimensional feature space. To address this challenge, latent space-based supervised learning has been developed to uncover latent dimensions of neural circuits in psychiatric disorders. For example, a sparse latent space regression algorithm tailored for EEG data was recently developed to identify antidepressant-responsive brain signatures in major depression \citep{wu2020electroencephalographic}. By jointly estimating the spatial filters and regression weights under a convex optimization framework, the ML model was able to successfully reveal treatment-predictive signatures in a low-dimensional latent space. To address comorbidities among psychiatric disorders, dimensional approaches have been developed using statistical models capable of discovering the complex linear relationship between high-dimensional datasets. For instance, low-dimensional representations of depression-related connectivity features have been successfully identified by applying canonical correlation analysis (CCA) to resting-state fMRI (rs-fMRI) connectivity and clinical symptoms \citep{drysdale2017resting}. The discovered representations defined two disease dimensions corresponding to an anhedonia-related component and an anxiety-related component, respectively. A similar dimensional analysis was also utilized to examine the neural correlates of neuropsychiatric symptoms in dementia. Using CCA, two latent modes were captured with the distinct neuroanatomical basis of common and mood-specific factors of the symptoms \citep{kwak2020multivariate}. A sparse CCA approach has been applied to reveal linked dimensions of psychopathology and functional connectivity in brain networks for psychiatric disorders \citep{xia2018linked}. This approach successfully identified interpretable dimensions, involving mood, psychosis, fear, and externalizing behavior, guided by neural circuit patterns across the clinical diagnostic spectrum. The partial least squares (PLS) approach was also applied to identify latent components linking a broad set of behavioral measures to functional connectivity \citep{kebets2019somatosensory}. The latent components defined distinct dimensions with dissociable brain functional signatures, thus providing potential intermediate phenotypes spanning diagnostic categories. These dimensional analytics hold great promise in uncovering novel transdiagnostic phenotypes for the development of targeted interventions.

\textbf{Ensemble Learning.} Though ML approaches have been extensively designed for supervised learning, using a single model may not produce the optimal generalization performance for a complex prediction task. By combining multiple ML models to reduce variance or bias, ensemble learning improves prediction performance over a single model and has proven successful in the robust discovery of biomarkers for psychiatric disorders. For instance, multi-atlas ensemble-learning algorithms have been proposed for improved schizophrenia detection \citep{kalmady2019towards} and ASD diagnosis \citep{khosla2019ensemble}. By utilizing multimodal neuroimaging including sMRI, fMRI, and DTI, a bagging-based SVM was devised to yield significant improvement in the prediction of adult outcomes in childhood-onset ADHD \citep{luo2020multimodal}. Based on the selective ensemble algorithm, a sparse multi-view prediction model has been designed with rs-fMRI connectivity for ASD diagnosis \citep{wang2020multi}. The model combined multiple classifiers under a bootstrap framework and significantly outperformed other single-model approaches.

Although sophisticated models of supervised learning often produce better classification or prediction performance, their interpretability decreases with the increasing model complexity. We will discuss the interpretable ML methods in more detail later (Section \ref{section-XAI}). Additionally, labeled data require the knowledge of the ground truth, which is not always accurate or reliable in the case of mental disorders. For instance, the skin cancer diagnosis may rely on training samples that have been biopsied and cataloged, leaving no doubt as to whether they are malignant or not; however, there is no equivalent of the biopsy in mental disorder.

\textbf{Unsupervised Learning.} Unsupervised learning relaxes the assumption of labeled samples and can be useful for exploratory data analysis. Unsupervised learning aims to uncover the intrinsic data structure by either identifying potential clusters (e.g., using latent class analysis or K-means clustering) or learning a feature mapping that satisfies certain criteria (e.g., using PCA). Identifying patient subtypes offers a promising strategy to delineate neurobiological heterogeneity in psychiatric disorders \citep{feczko2019heterogeneity}. With rs-fMRI, hierarchical clustering was applied to successfully identify four subtypes of functional connectivity in depression \citep{drysdale2017resting}. These subtypes were found to correlate with differing clinical-symptom profiles and predict responsiveness to brain stimulation therapy. From rs-EEG, two transdiagnostic subtypes were identified using sparse K-means clustering with distinct power envelope connectivity patterns and found to respond differentially to antidepressant medication and psychotherapy \citep{zhang2021identification}. As a non-distance probability-based clustering approach, latent class analysis has also been applied to discover subgroups in psychiatric disorders. A proof-of-concept study was conducted using latent class analysis to identify ADHD subtypes from fMRI activation profiles \citep{lecei2019can} and reveal that the subtype with attenuated brain activity showed fewer behavior problems in daily life. By leveraging data resources from multiple time points, psychiatric studies have been shifting from cross-sectional analysis to longitudinal modeling \citep{etkin2019reckoning}. Finite mixture modeling became increasingly popular for the analysis of longitudinally repeated-measure data, which can identify latent classes following similar paths of temporal development \citep{elmer2018using,van2020overview}. Typical finite mixture models include growth mixture modeling, group-based trajectory modeling, and latent transition analysis. The use of latent growth mixture modeling (LGMM) and group-based trajectory modeling has been growing in studying psychiatric disorders, such as depression, anxiety, and ASD. They offer flexible ways to identify latent subpopulations that manifest heterogeneous symptom trajectories \citep{ellis2022latent,ulvenes2022latent,waizbard2022identifying}. LGMM approaches have also been successfully used to predict PTSD course among the population at risk \citep{schultebraucks2020natmed}. As an extension of latent class analysis to longitudinal data, latent transition analysis has been applied to predict longitudinal service use for individuals with substance use disorder \citep{crable2022predicting}. Together, these approaches provide powerful tools to delineate longitudinal heterogeneity and the corresponding distinctive phenotypes during the course of psychiatric disorders.

\textbf{Semi-supervised Learning.} Semi-supervised learning is an ML approach that combines supervised learning and unsupervised learning. Popular semi-supervised learning techniques include self-training, mixture models, co-training and multi-view learning, graph-based methods, and semi-supervised clustering \citep{chapelle2009book}. These methods have been increasingly applied to psychiatric studies. By unifying autoencoder and classification, a semi-supervised model was developed for ASD diagnosis \citep{yin2022semi}. A semi-supervised classification has been devised using graph convolutional networks and applied to the population graph-based diagnosis of ASD \citep{parisot2018disease}. A semi-supervised clustering has also been designed by extending SVM with implicit clustering driven by a convex polytope to form a method called heterogeneity through discriminative analysis, which can achieve joint disease subtyping and diagnosis \citep{varol2017hydra}. This approach has shown strength in delineating neurostructural heterogeneity in bipolar and major depressive disorders (MDDs) \citep{yang2021probing}, schizophrenia \citep{honnorat2019neuroanatomical}, as well as in youth with internalizing symptoms \citep{kaczkurkin2020neurostructural}. Additionally, semi-supervised learning has gained increasing mental health applications in digital data from electronic health records (EHRs), social media and mobile phones \citep{beaulieu2016semi,yazdavar2017semi,dong2021semi}. See Section \ref{section-ML_Psychiatry} for a detailed discussion.

Semi-supervised learning also provides an elegant way through normative modeling \citep{marquand2016understanding} to characterize neurobiological heterogeneity by quantifying individual deviations. By building a normative model of neuroimaging data on a large-scale healthy population, brain abnormalities of individual patients can be quantified by examining their statistical differences from the distribution of the norm. Gaussian process (GP) regression-based normative modeling has been applied to quantify individual deviations and dissect neurobiological heterogeneity in various psychiatric disorders \citep{rutherford2021normative}. With this tool, an association was successfully discovered between transdiagnostic dimensions of psychopathology and individual’s unique deviations from normative neurodevelopment in brain structure \citep{parkes2021transdiagnostic}. By combining tolerance interval-based normative modeling and clustering analysis, individual abnormalities in rs-fMRI were accurately quantified to define two stable subtypes in patients with PTSD \citep{maron2020individual}. The two subtypes showed distinct patterns of functional connectivity with respect to the healthy population and differed clinically on levels of reexperiencing symptoms. These novel data-driven approaches provide useful techniques to identify ``abnormal" subtypes in patients, thereby advancing clinical and mechanistic investigations in psychiatric disorders.

\begin{figure}[!t]
\centering
\includegraphics[width=9cm]{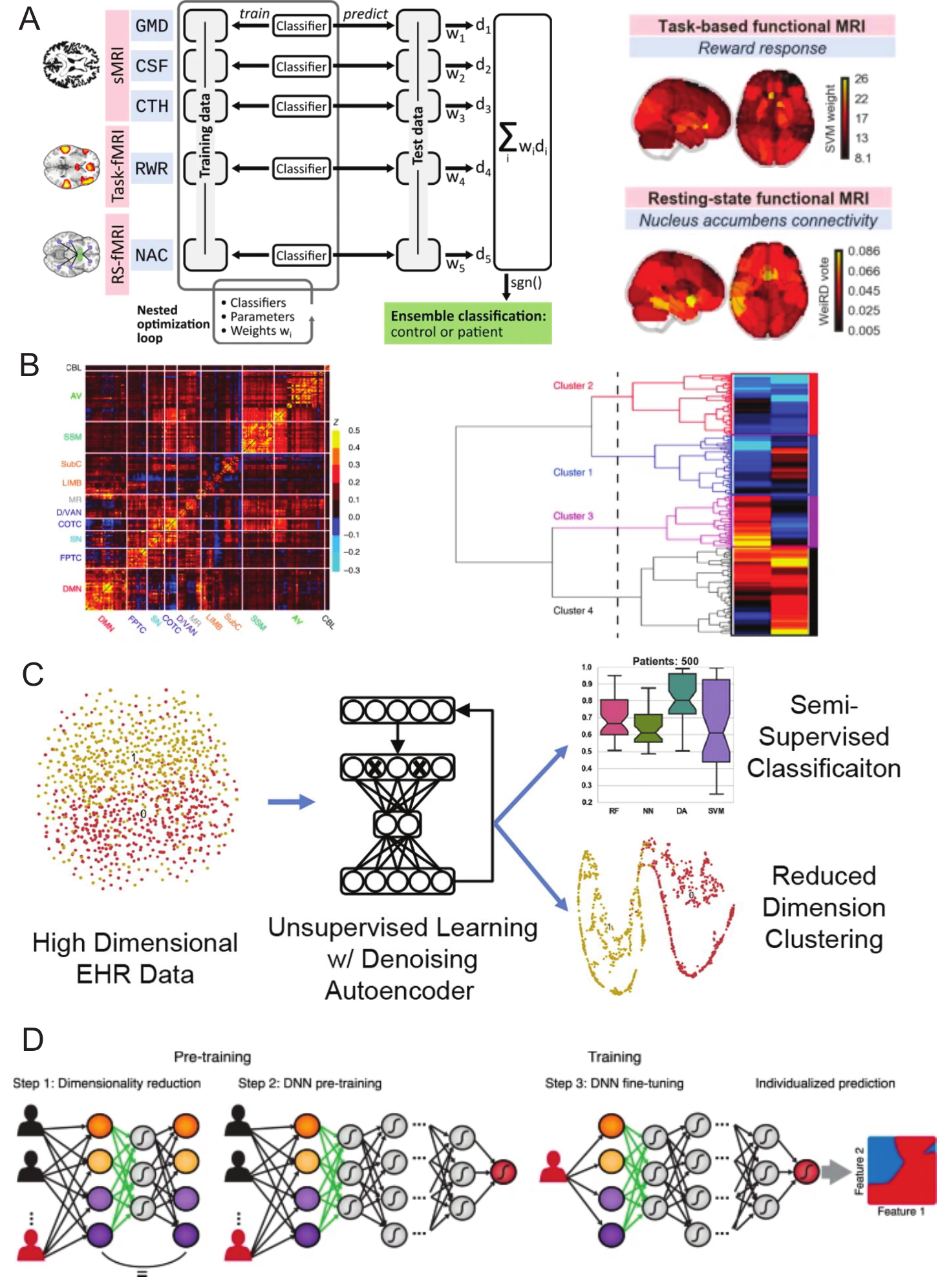}
\caption{Various ML models for mental health applications.
(A)	Left: Multimodal supervised classification scheme. Three modality-specific factors are optimized on the training data: classifier types, parameters and weights. The final diagnostic classification is based on a weighted sum of decision values, where weights correspond to those estimated during training. Right: Feature importance maps of functional neuroimaging modalities (\citep{guggenmos2020multimodal}). 
(B)	Unsupervised learning. Left: whole-brain functional-connectivity matrix averaged across all subjects. z = Fischer transformed correlation coefficient. Right: Hierarchical clustering analysis. ($\copyright$ Springer Nature, \citep{drysdale2017resting}. Figure reproduced with permission).
(C)	Semi-supervised learning pipeline for phenotype stratification based on EHRs (\citep{BEAULIEUJONES2016}, Figure reproduced with permission)
(D)	Deep neural networks (DNNs) for group-level and individualized treatment predictions. Future data points could then be used to forecast symptom onset, treatment response, or other mental health-related variables (\citep{koppe2021deep}, Creative Commons licenses 4.0).}
\label{fig2}
\end{figure}

\subsection{Deep Learning}
Deep learning consists of a collection of methods that use artiﬁcial neural networks for machine learning tasks. Through a specifically designed deep neural network structure, high-level feature representations can be learned from raw features. Deep learning thus holds promise in offering an end-to-end analytic framework for disease diagnosis and prediction. With the advancement in neuroimaging technologies, an increasing number of large-scale multi-center datasets have been established for building powerful ML models to fully explore the informative feature representations from the complex brain and genomic data. By training on these large-scale datasets, deep learning can learn robust neuroimaging representations and outperform standard ML methods in a variety of application scenarios in mental health \citep{shen2017deep,abrol2021deep,koppe2021deep,quaak2021deep}.

\textbf{Deep autoencoder.} The deep autoencoder, also known as stacked autoencoder, aims to learn latent representations of input data through an encoder and uses these representations to reconstruct output data through a decoder. By stacking multiple layers of autoencoders, deep autoencoder is formed to discover more complicated and potentially nonlinear feature patterns. Deep autoencoder has been applied to extract low-dimensional features from the amplitude of low-frequency fluctuations in fMRI \citep{chang2021identifying}. Clustering analysis with the latent features uncovered by deep autoencoder further identified two subtypes within major psychiatric disorders including schizophrenia, bipolar disorder, and MDD. A deep learning model was also designed based on a sparse stacked autoencoder and applied to lower the dimensionality of fMRI connectivity. The sparsity constraint used in this model yielded interpretable neural patterns for improved ASD diagnosis \citep{almuqhim2021asd}. Deep autoencoder has also been applied to implement normative modeling with structural MRI for the quantification of individual abnormalities in neuropsychiatric disorders, including schizophrenia and ASD \citep{pinaya2019using}. The abnormal features extracted using the normative model led to improved diagnosis performance compared with the traditional case-control analysis. Recently, a deep contrast variational autoencoder was used to extract neuroanatomical features from MRI data to identify brain dysfunction that can be attributed to ASD and not to other causes of individual variation \cite{Aglinskas22_sci}.

\textbf{Convolutional neural networks (CNNs).} Different from conventional multi-layer perceptron or autoencoder assigning a different weight to each input feature, CNNs were designed to better capture the spatial and local structure information from pixels or voxels \citep{anwar2018medical,zhang2020survey}. Due to its strength in utilizing neighborhood information to learn hierarchies of features \citep{yamashita2018convolutional}, CNN has been one of the most successful deep learning models applied in various medical applications. A diagnosis model was established through EEG-based image construction coupled with the CNN for accurate detection of MDD \citep{uyulan2021major}. This model provided an end-to-end framework to successfully identify translational biomarkers from resting-state EEG in distinguishing depressive patients from healthy people. With whole-brain structure MRI, a 3D CNN model has also been designed to automatically extract multilayer high-dimensional features for the diagnosis of conduct disorder \citep{zhang2020three}.

\textbf{Graph neural networks (GNNs).} Though deep learning models have shown strengths in capturing complex neuroimaging patterns, they may not generalize well to non-Euclidean data types (e.g., brain networks). In contrast, GNNs provide a clever way of learning the deep graph structure of non-Euclidean data, leading to enhanced performance in various network neuroscience tasks \citep{bessadok2021graph}. For instance, a framework based on graph convolutional networks has been designed for the diagnosis of ASD \citep{parisot2018disease}. By building a population graph that integrates rs-fMRI data as node features and phenotypic measures as edges, the designed model outperformed other state-of-the-art methods. An inductive GNN model was also devised to embed the graphs containing different properties of task fMRI and drive interpretable connectome biomarkers for ASD detection \citep{li2021braingnn}. More recently, a novel GNN model was developed to incorporate dynamic graph computation and feature aggregation of 2-hop neighbor nodes into graph convolution for brain network modeling \citep{zhao2022dynamic}. This dynamic GNN significantly improved the performance in ADHD diagnosis and revealed the circuit-level association between connectomic abnormalities and symptom severity.

\textbf{Recurrent neural networks (RNNs).} As a specific extension of the feed-forward neural network, RNN has the ability to learn features and long-term dependencies from sequential and time-series data. Long-short-term memory (LSTM) model is the most popular RNN and has shown its advantage in capturing temporal dynamic information of neuroimaging data for various psychiatric disorder studies \citep{durstewitz2021psychiatric}. An LSTM-based RNN architecture was built with the time course of fMRI-independent components to exploit the temporal information, which yielded an improved diagnosis of schizophrenia \citep{yan2019discriminating}. By combining RNN with other deep neural networks, novel machine learning models have also been proposed to model the spatio-temporal dynamics in neuroimaging data. A spatio-temporal CNN model was proposed for 4D modeling of fMRI, with confirmed robustness in identifying key features in the default mode network \citep{zhao20204d}. LSTM has also been applied to incorporate multi-stage neuroimaging data into longitudinal analytic frameworks for modeling the trajectories of psychopathology development in various psychiatric disorders. A recent LSTM-based model was built with MEG data to achieve accurate longitudinal tracking of pathological brain states and prediction of clinical outcomes in PTSD \citep{zhang2020predicting}.

\textbf{Generative Adversarial Networks (GANs).} GAN is a type of generative model, which has gained considerable attention in computer vision and natural language processing and also become increasingly popular in neuroimaging analysis \citep{zhang2020survey}. GAN consists of two competing neural networks (one as generator and the other as discriminator) and can learn deep feature representations without extensive labeled data. Due to this unique advantage, GAN has been increasingly applied in data augmentation to enhance the sample size for model training \citep{lashgari2020data}. Moreover, GAN has been used to impute missing values in multimodal datasets, a common problem in psychiatric studies, rather than discarding an entire multivariate data point \citep{shang2017vigan}. The adversarial model has also been incorporated into other ML models for specific applications in psychiatric studies. For instance, the discriminative and generative components were incorporated in LSTM to form a multitask learning approach for fMRI-based classification, which resulted in an improved diagnosis of ASD compared with the standard LSTM  \citep{dvornek2019jointly}. By integrating GAN with group ICA, a functional connectivity-based deep learning model was developed for the diagnosis of MDD and schizophrenia \citep{zhao2020functional}. Specifically, the generator with fake connectivity was trained to match the discriminator with real connectivity in the intermediate layers, whereas a new objective loss was determined for the generator to improve the diagnosis accuracy.

The strength of deep learning algorithms is that they can learn complex predictor-response mappings, but the power also comes at the cost of requiring a very large sample size for model optimization. This poses potential overfitting and interpretability challenges in psychiatric applications \citep{koppe2021deep}.

\begin{figure*}[t]
	\centering
	\includegraphics[width=0.9\textwidth]{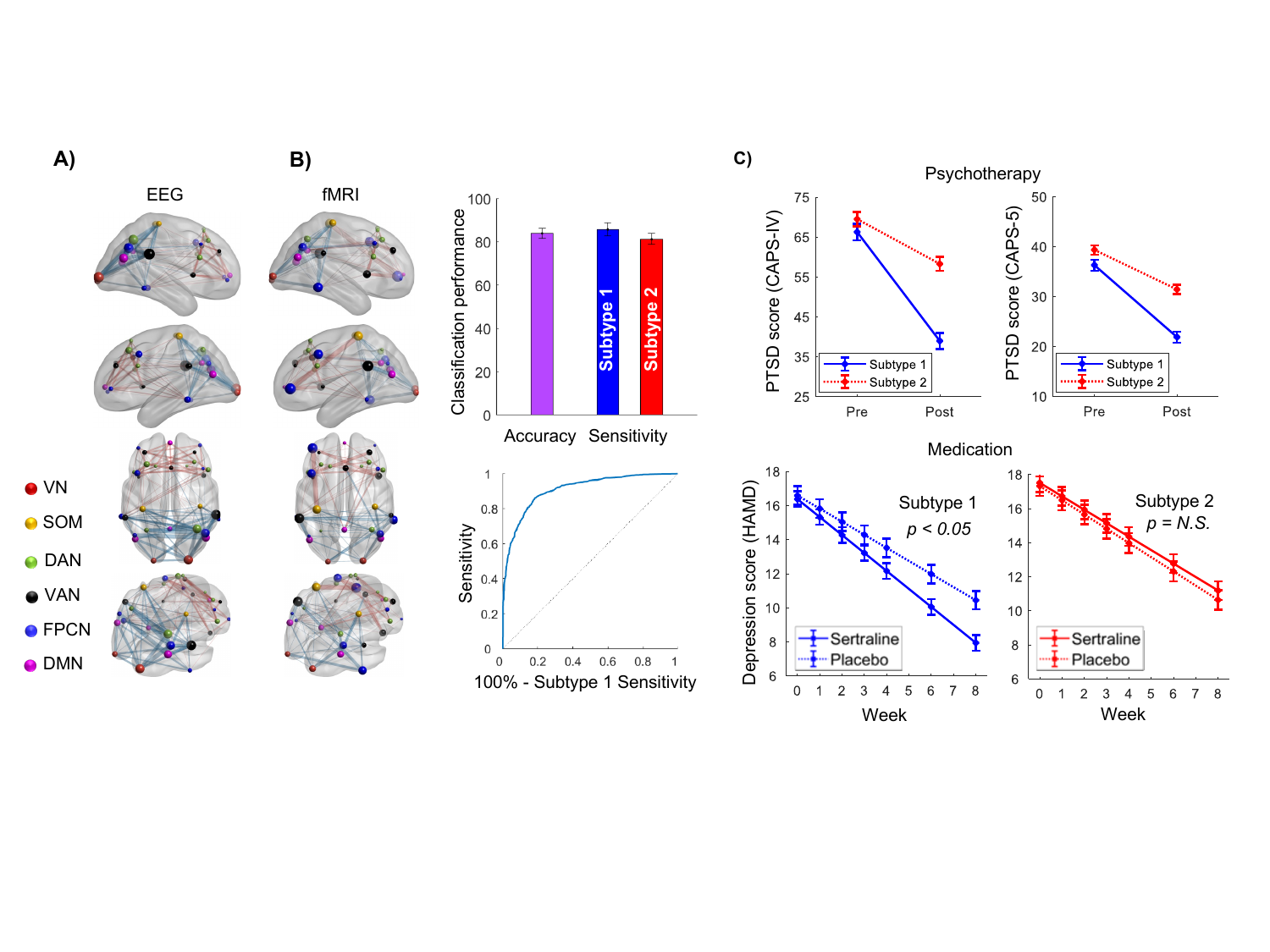}
	\caption{Concepts and major findings in case studies 1 and 2. (A) Illustration of the sparse EEG latent space regression (SELSER) framework in Case Study 1 for treatment outcome prediction. (B) Interpretable cortical pattern derived from the scalp pattern ($\copyright$ Springer Nature, figures are modified from \citep{wu2020electroencephalographic} with permission). (C) Distinctive EEG connectivity profiles were identified by sparse K-means for defining psychiatric subtypes in Case Study 2 on PTSD and MDD. The two identified subtypes were further found to predict treatment responsiveness to psychotherapy and antidepressant medication. (D) The EEG connectivity-defined subtypes are distinguishable by rs-fMRI connectivity patterns derived from an RVM-based classifier ($\copyright$ Springer Nature, figures are modified from \citep{zhang2021identification} with permission).} 
	\label{Fig-CaseStudy12}
\end{figure*}

\subsection{Key ML Concepts for Precision Psychiatry}
Regardless of the ML paradigms in psychiatric applications, there are some common themes that distinguish between human intelligence and automated or human-in-the-loop machine intelligence. In a recently published white paper ``{\it Machine intelligence for healthcare}", four important features are emphasized for ML systems \citep{cutillo2020machine}. These concepts are broadly applicable to precision psychiatry \citep{chandler2020using}.

\begin{itemize}
\item \textbf{Trustworthiness}: the ability to access the validity and reliability of an ML-derived output across varying inputs and environments. In other words, psychiatrists need to be able to evaluate the limitations of an ML system and confidently apply system-derived information for psychiatric evaluation.
\item \textbf{Explainability}: the ability to understand and evaluate the internal mechanism of a machine. The development of ML systems will need to account for data quality, quality metrics for the system’s functioning and impact, standards for applications in the environment, and future updates to the system. 
\item \textbf{Usability}: the extent to which an ML system can be used to achieve specified goals with effectiveness, efficiency, and patient satisfaction in multiple environments. These applications need to be scalable across multiple settings while preventing additional burdens on providers and patients.
\item \textbf{Transparency and Fairness}: the right to know and understand the aspects of an input that could influence outputs (clinical decision support) from the system. Such factors should be available to the people who use, regulate, and are affected by any type of care decision that employs the ML system. The potential bias in the data needs to be identified and informed prior to decision making.
\end{itemize}

The first two features are related to interpretability, which we will discuss in more detail in Section \ref{section-XAI}. The other two features will be discussed in Section \ref{section-Discussion-Conclusion}.

\begin{figure*}
	\centering
	\includegraphics[width=0.9\textwidth]{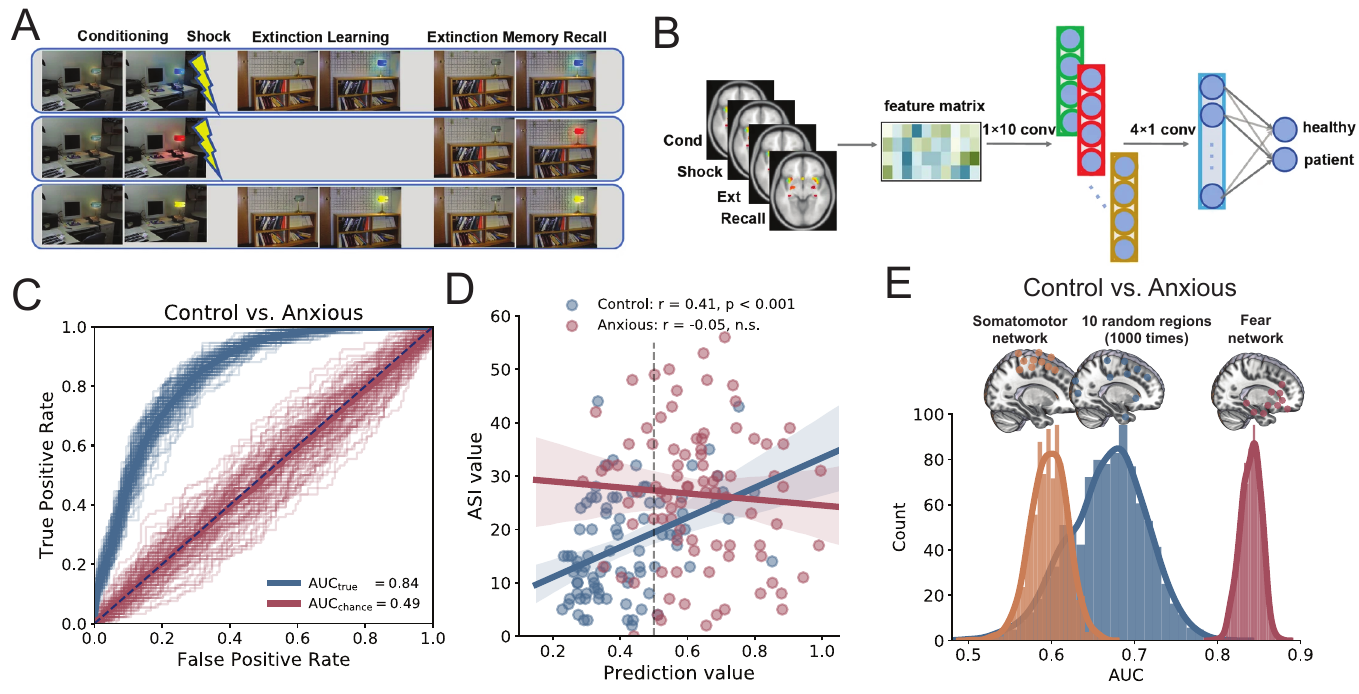}
	\caption{Illustrations of concepts and major findings in case study 3. (A) Experimental paradigm. (B) Schematic of the CNN. (C) AUC curves produced by CNN vs. chance level. (D) The prediction score positively correlated with the anxiety sensitivity index (ASI) for the control group ($r=0.41$, $p<0.001$), but at the chance level for anxious brains ($r=-0.05$, $p=0.65$). (E) Distribution of AUCs based on brain activations within the 10-node fear randomly selected brain regions (Figures were adapted from \citep{wen2021fear} with permission).}
	\label{CaseStudy3}
\end{figure*}

\begin{figure}[!t]
	\centering
	\includegraphics[width=9cm]{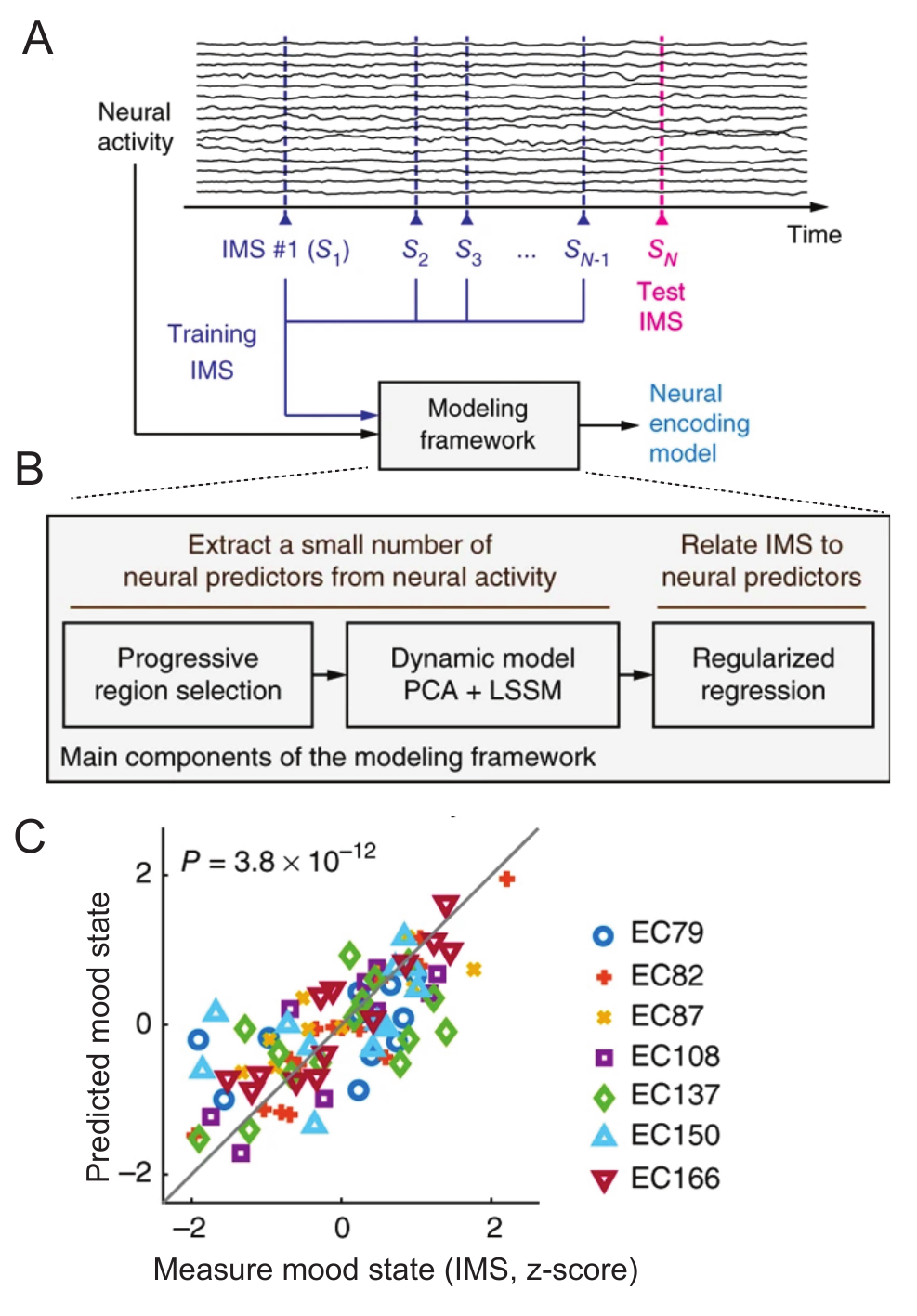}
	\caption{Illustrations of concepts and major findings in case study 4. (A) Schematic of cross-validation. An IMS point (e.g., $S_N$) is left out as the test IMS to be predicted. The other IMS points (i.e., training IMS, using $S_1$ to $S_{N–1}$) and the associated neural activity are used within the modeling framework to train a neural encoding model. (B) Main components of the modeling framework based on both unsupervised and supervised learning. (C) Cross-validated prediction of the mood state is shown against the true measured mood state ($\copyright$ Springer Nature; Figures were modified from \citep{sani2018mood} with permission).}
	\label{CaseStudy4}
\end{figure}

\subsection{Case Studies}
To help reader get a concrete idea of the reviewed ML techniques in psychiatric applications, here we present several case studies to illustrate the strengths in prediction/classification diagnosis analytics. These representative case studies employ different ML strategies and cover different data modalities, including rs-EEG, task fMRI, and ECoG.

{\it Case Study 1: Sparse latent space learning for EEG-based treatment prediction in depression.} Antidepressants have shown only modest superiority over placebo, which is partly because the clinical diagnosis of MDD encompasses biologically heterogeneous conditions that relate differentially to treatment outcomes. A robust neurobiological signature for an antidepressant-responsive phenotype is still lacking to determine which patients will benefit from medications. To address the challenge, Wu et al. \citep{wu2020electroencephalographic} developed a sparse EEG latent space regression (SELSER) model to predict the treatment outcome. Specifically, SELSER optimizes the spatial filters and regression weights in conjunction under a convex optimization framework, and identifies an antidepressant-responsive EEG signature for MDD  (Figure~\ref{Fig-CaseStudy12}A). The identified signature accurately predicts antidepressant outcomes ($n=228$). A neurophysiologically interpretable cortical pattern was further observed through a source mapping from the scalp spatial pattern, mainly contributed by the right parietal-occipital regions and the lateral prefrontal regions (Figure~\ref{Fig-CaseStudy12}B). The validation on an independent cohort showed that the treatment outcomes predicted by the brain signature are significantly higher in a partial responder group versus a treatment-resistant group, demonstrating its further clinical utility in the broader construct of treatment resistance in depression.

{\it Case Study 2: Unsupervised learning-based identification of neurophysiological subtypes in psychiatric disorders.} Neurobiological heterogeneity has a substantial impact on treatment outcome independent of pre-treatment clinical symptoms. For example, although psychotherapy is currently the most effective treatment for PTSD, many patients are nonetheless non-responsive and display differences in brain function relative to responsive patients. Using sparse K-means clustering, Zhang et al. \citep{zhang2021identification} developed a data-driven framework to achieve simultaneous feature selection and subtyping on the high-dimensional power envelope connectivity of rs-EEG source-reconstructed signals. This approach successfully identified two transdiagnostic subtypes with distinct functional connectivity patterns in PTSD and MDD ($n=648$), prominently within the frontoparietal control network and default mode network (Figure~\ref{Fig-CaseStudy12}C). Importantly, linear mixed models in an intent-to-treat analysis on symptom severity revealed that the two subtypes differentially responded to psychotherapy and antidepressant versus placebo. An RVM-based classification analysis further confirmed that the EEG connectivity-driven subtypes were distinguishable using rs-fMRI connectivity. The discriminative pattern identified from fMRI was also consistent with the EEG connectivity pattern (Figure~\ref{Fig-CaseStudy12}D).

{\it Case Study 3: Classification of anxious vs. non-anxious brains from fear extinction learning task-based fMRI.} Using a neuroimaging cohort study ($n=304$ adults, 92 anxiety patients, 74 trauma-exposed individuals, 138 matched controls), Wen et al. \citep{wen2021fear} examined how the fMRI activations of 10 brain regions that were commonly activated during fear conditioning and extinction (Figure~\ref{CaseStudy3}A) might distinguish anxious or trauma-exposed brains from controls. They proposed a CNN classifier (Figure~\ref{CaseStudy3}B) to map fear-induced fMRI activities in space and time to a prediction probability score indicating that the subject belongs to the anxious group. The CNN achieved an AUC of $0.84 \pm 0.01$, $0.75 \pm 0.03$ sensitivity, and $0.77 \pm 0.02$ specificity in 5-fold cross-validation (Figure~\ref{CaseStudy3}C), outperforming other ML methods (e.g., SVM and random forest). The prediction score was also found to correlate with the anxiety sensitivity index (ASI) in the control group (Figure~\ref{CaseStudy3}D). Furthermore, control analyses were performed to demonstrate the specificity of the fear network in discrimination (Figure~\ref{CaseStudy3}E).

{\it Case Study 4: Decoding mood state from multi-site intracranial brain activity.} From intracranial ECoG signals and simultaneously collected self-reported mood state measurements over multiple days in seven epilepsy patients, Sani et al.  \citep{sani2018mood} developed a dynamic state-space model (SSM) framework to track the patients' mood state variations over time (Figure~\ref{CaseStudy4}A). The modeling framework consists of unsupervised and supervised learning components (Figure~\ref{CaseStudy4}B). The spectro-spatial features were extracted from the mood-predictive network within the limbic brain region. The neural decoders were also highly predictive of the immediate mood scaler (IMS) points at the population level. Furthermore, the same trained decoder could be used for mood state prediction across hours and days, and generalized across a wide range of IMS. In cross validation, the decoders could predict IMS variations that covered 73\% and $33 \pm 7.2$\% of the total possible IMS range across all seven subjects and within individuals, respectively (Figure~\ref{CaseStudy4}C). These results suggest that ML-based decoders can predict mood state variations from brain activity across multiple days of recordings in patients.

\section{ML-powered Technologies for Psychiatry}
\label{section-ML_Psychiatry}
ML can be applied to a wide range of digital platforms, including software (e.g., mobile apps), hardware (e.g., wearable devices, robots), social services (e.g., online chatbots) and clinical practice (e.g., EHRs). In this section, we will review various ML-powered technologies in the non-neuroimaging domains and highlight the emerging digital platforms and their underlying ML technologies for precision psychiatry.

A recent McKinsey study  showed that use of telehealth has increased by 38-fold as compared to the pre-COVID baseline \citep{Bestsennyy2021}. With a steep increase in teletherapy demand and consumption, many companies (such as Talkspace  and Headspace Health) provide services on chat conversations with licensed mental health professionals. The definition of teletherapy has expanded to include these newer modalities of care delivery. These advances in care delivery have enabled collecting massive amounts of text, audio and video data on a regular basis, which was previously only available in controlled research settings. Furthermore, the recent advancements of natural language, speech, and video analysis technologies, combined with the ML tools, have generated numerous innovations in the emerging field. The global psychiatrist community is increasingly aware of these developments. For example, a recent survey among more than 700 psychiatrists showed that 49\% believed that in the next 5-10 years, ML technology will help analyze patient information to establish prognosis and 54\% believed that this technology can help synthesize patient information to reach a diagnosis \citep{doraiswamy2020artificial}.

ML can be applied  across all stages of a patient’s journey \citep{lee2021artificial,moustafa2021big}: risk assessment, diagnosis, prognosis, treatment, and remission in a variety of disorders \citep{graham2019artificial}, where the analytics can be applied to  natural language, speech, facial expressions, body language, social media, as well as traditional clinical surveys and neuroimaging data \citep{su2020deep,abbas2021digital}. Table~\ref{tab4} summarizes  recent representative studies that use ML to support various stages of patient journey. Applying ML can build personalized models  that are optimized for each patient \citep{bzdok2018machine}, as opposed to traditional models that are only optimized for group effects.  Furthermore, given the inter and intra-disease variability between clinical diagnosis and symptoms, ML can be used to model the differential diagnosis between disease categories using methods like multi-task learning. All of these mentioned ML applications  can be considered to be the first level of precision added to ML-powered psychiatry.

\begin{figure*}[t]
	\centering
\includegraphics[width=0.9\textwidth]{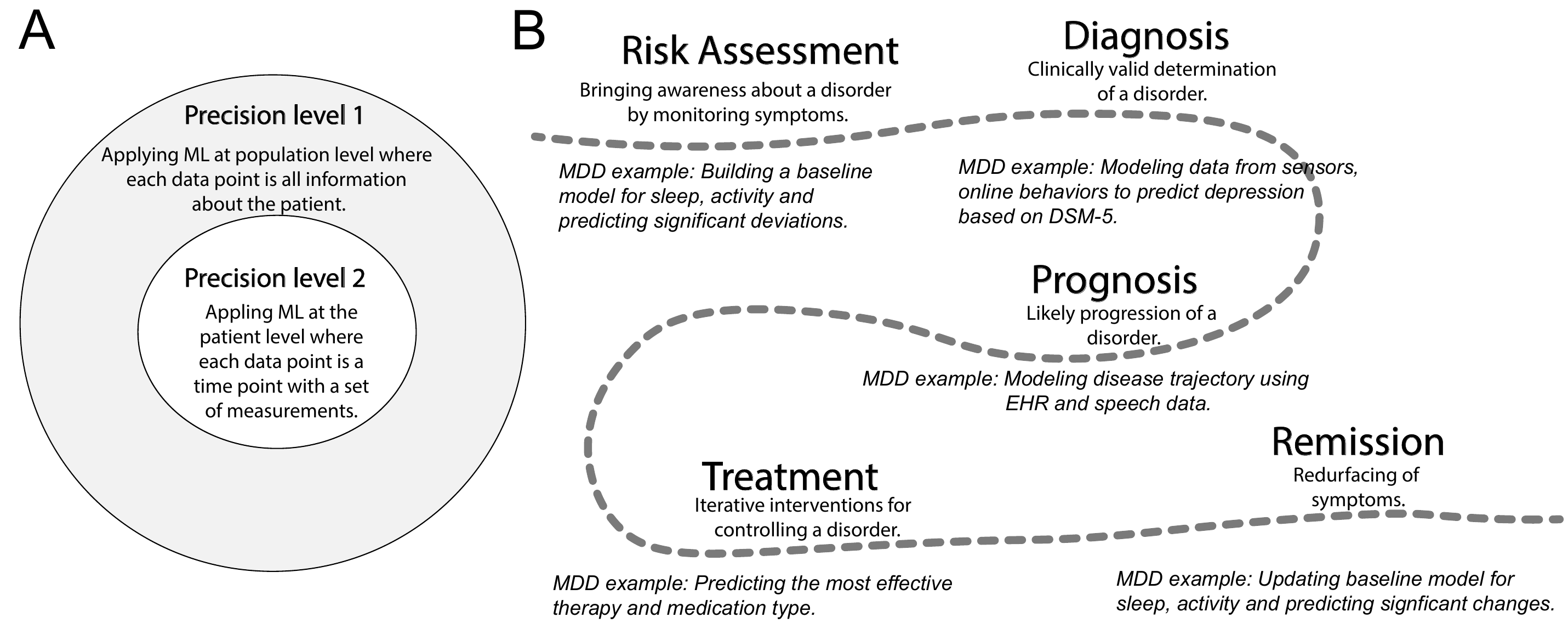}
\caption{(A) Two levels of precision in applying ML for mental health. (B) Examples of ML applications at various stages of a patient’s journey in case of major depressive disorder (MDD). }
\label{fig_MDD}
\end{figure*}

However, the amount of precision  that can be modeled using ML is far beyond the first level  \citep{bickman2020improving,wilkinson2020time}. During psychiatric evaluation, psychiatrists may try to build a mental model of what is going on in the patient’s life in about 30 minutes. They aim to understand as much as possible about the patient’s history in a very short time,  define what ``normal" looks like for the patient,  and identify deviations from that normal. This is often done by asking the patient questions and examining their speech, body language, and behavioral responses. It is very challenging and almost unrealistic to expect psychiatrists to build an accurate baseline model of the patient’s entire life in such a short time span whilst interacting with the patient in a compromised psychological state. ML can help by building baseline models specific to each patient before their visit and present the bounds for various observations as a reference to psychiatrists during the exam \citep{barron2021reading}. This can be viewed  the second level of precision in psychiatry that can be made possible by ML (Figure~\ref{fig_MDD}A).  Take MDD as an example, Figure~\ref{fig_MDD}B shows how ML can be applied at different stages of a patient's journey. Similar applications have also been developed in studies of other disorders. 

In the following subsections, we describe how ML technologies can be applied to clinically relevant data and to support one or more stages of the patient's journey.

\begin{table*}
\centering
    \caption{ Representative ML applications of multimedia data in mental disorders.}
   \scalebox{0.8}{
  \begin{tabular}{lllll}
  \hline
\bf Study &	\bf Data source(s) &	\bf Patient journey stage &	\bf ML approach	& \bf Test sample size    \\
  \hline
  
 \multicolumn{4}{c}{ Depression spectrum } \\ \hline 
  
\citep{vazquez2020automatic} &	Audio - clinical interviews	& Diagnosis	& CNN ensemble & $47$ speakers \\

\citep{harati2021speech} &	Audio - answers to personal questions  & 	Diagnosis &	Transfer learning 	& $3078$ speakers \\
 
\citep{huang2019investigation} &	Audio - clinical interviews	& Diagnosis &	SVM w/ speech landmark features	& $47$ speakers \\

\citep{zhu2017automated}	& Face video -reading \& personal questions  &	Diagnosis &	CNN	& $50$ videos \\

\citep{shao2021multi} &	Gait-only video - casual walking in a corridor &	Diagnosis &	LSTM+CNN weighted fusion	& $40$ videos \\

\citep{lu2020robust} &	Language - answers to personal questions  & 	Diagnosis &	LSTM fine-tuned w/ health forum data & $2425$ subjects \\

\citep{eichstaedt2018facebook} &	Language - Facebook posts &	Risk assessment	& Logistic regression &	$68$ patients \\

\citep{sun2021multi}	& Audio, video - clinical interviews &	Diagnosis &	Transformer + multimodal fusion & $56$	subjects \\ \hline 

 \multicolumn{4}{c}{ Bipolar spectrum } \\ \hline 
 
\citep{weiner2021vocal}	& Audio - verbal fluency tasks &	Remission &	SVM	& 56 subjects \\ 
\citep{palmius2016detecting} & Sensor - GPS	& Diagnosis &	Linear regression 	& 36 subjects \\ \hline

  \multicolumn{4}{c}{ PTSD } \\ \hline 
 
 \citep{marmar2019speech}	& Audio - clinical interviews &	Diagnosis	& Random forest &	43 veterans \\
 
\citep{mallol2018multimodal} &	Audio, video, skin conductance  &	Remission	& SVM & 110 subjects  \\ \hline

 \multicolumn{4}{c}{ Schizophrenia  spectrum } \\ \hline 

\citep{tahir2019non} &	Audio - clinical interviews	& Diagnosis &	SVM	& 70 subjects \\

\citep{abbas2021computer} &	Video - neutral open-ended questions &	Diagnosis &	Logistic regression &	16 subjects \\

\citep{birnbaum2020utilizing} &	Language - internet search queries &	Remission	& Random forest	& 23 subjects \\

\citep{birnbaum2022acoustic} &	Audio, video - clinical interviews &	Diagnosis	& Gradient boosting &	17 subjects \\

  \hline
\end{tabular}}
\label{tab4}
\end{table*}

\subsection{Mobile and Sensing Technologies}
\begin{figure}[!t]
\centering
\includegraphics[width=9cm]{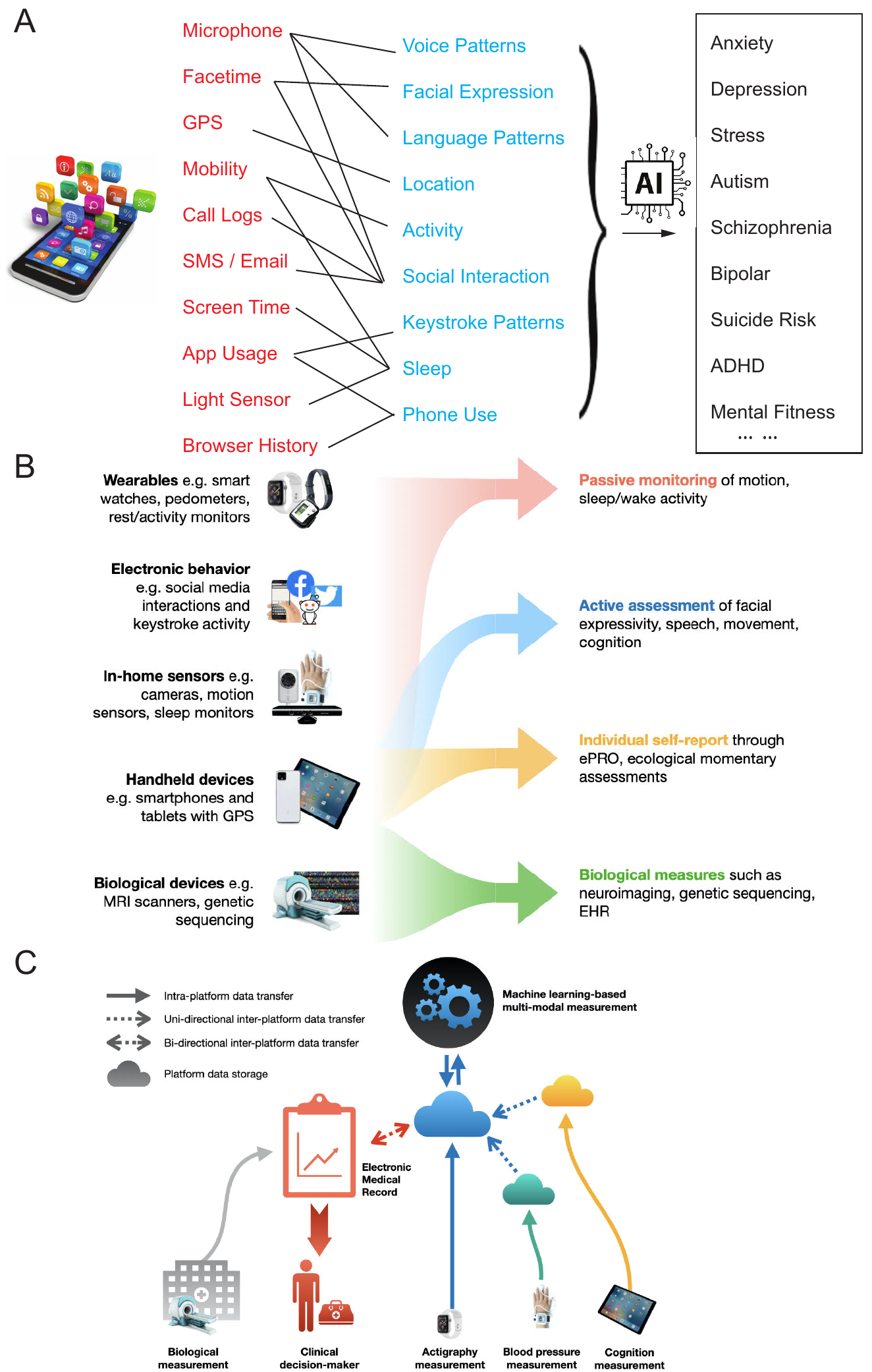}
\caption{Illustrations of ML-powered technologies for mental health. (A) ML applications in mobile health. (B) Different types of data collection strategies for digital measurement tools. (C)  A technological infrastructure for the integration of digital measurement tools. Independent
platforms for measurement of health will have their own
data repositories, depicted as clouds. This data could be safely transferred across platforms using transfer tools such as secure APIs (application program interfaces), depicted using dashed arrows. Such tools could allow for both unidirectional and bidirectional movement of data. ML can be applied to integrate all measures for clinical decision-making (panels B and C are reproduced from \citep{abbas2021digital} with permission).}
\label{fig_tech}
\end{figure}

The development of smart phones, smart watches and other wearable sensing devices have enabled us to access more information of our physical and mental health than ever \citep{abdullah2018sensing}. Specifically, several types of signals are relevant for mental health monitoring and assessment (Figure~\ref{fig_tech}A):

\begin{itemize}
\item Behavioral and physical signals: location (e.g., GPS coordinates), mobility (e.g., accelerometer) 
\item Multimedia signals: face expression, speech patterns 
\item Social signals: social interactions (e.g., call and text message logs), communication patterns, engagement, online gaming
\item Physiological signals: skin conductance, hear rate variability (HRV), eye movement, electrodermal activity (EDA) 
\item Sleep activity: phone on/off status, sleep duration, sleep staging
\end{itemize}

These signals have different implications and relevance to mental illnesses. Although none of single signals is indicative of mental disorders, combination of these physical/physiological/social cues may reveal important clues of individual mental health. In what follows, we will focus on the analysis of multimedia, language, and social media data and on their mental health applications.

\subsection{Speech and Video Analyses}
Voice and visual (video of facial expressions and body language behaviors) data have recently gained increasing attention in the studies of mental disorders. ML technologies using speech samples obtained in the clinic or accessed remotely may help identify biomarkers to improve diagnosis and treatment. In the early stage of practice, psychologists have already used auditory and visual cues  to  assist the mental illness diagnosis \citep{kraepelin1921manic}. Furthermore, speech and video are  not only the readily available in traditional teletherapy settings but also the most interpretable as the most natural form of human communication.

\textbf{Audio and speech features.} Acoustic features derived from audio data have been found to be relevant in many mental health disorders \citep{marmar2019speech,low2020automated}, including speech analysis in patients with  depression, bipolar, and schizophrenia. Table~\ref{tab_feature} lists some commonly used acoustic features in the analysis of mental illnesses \citep{eyben2015geneva}. These categories have enabled standardization and  interpretation of ML-analyzed   speech data in clinical applications.

It has been shown that models built from speech-based features are effective in predicting the diagnosis of depression and suicidality \citep{cummins2015review}. Applications for depression include predicting the presence, severity, or score  \citep{vazquez2020automatic,weiner2021vocal}. These models use prosodic, spectral or other  features computed from raw speech data to quantify flattered speech, slow speech and other relevant markers. The target outcome variable is derived from a clinically valid scale such as a patient health questionnaire (PHQ-9). Furthermore, models for suicidality that explore similar features have been used in multi-class settings to differentiate between healthy, depressed, and suicidal speech. 

One key challenge in applying speech-based models in clinical practice is the lack of longitudinal validation of data acquired in real-world settings. However, this issue is starting to get addressed in recent studies  \citep{karam2014ecologically}, which detect manic and depressive speech from recordings of outgoing speech from phone conversations of consenting participants. Another remaining challenge is the lack of large labeled datasets for evaluating performance across various methods. To this end, it is noted that recent efforts  backed by companies like Ellipsis Health \citep{rutowski2021cross,harati2021speech}, have used deep learning and transfer learning to predict depression and anxiety scores with high accuracy based on a large labelled dataset of over 10,000 unique speakers. Human-level accuracy in detecting depression using only 20-30 seconds of audio clip has been reported in some commercial applications \citep{AIVoiceDepression,Sonde}.

\textbf{Visual features.} Although body language and facial expressions have always formed a key part of a psychiatric exam, ML has only recently been applied to analyze such data objectively. To date, most work  has been targeted to  suicidal ideation  \citep{galatzer2021validation},  depression \citep{zhu2017automated,song2018human,shao2021multi},  schizophrenia \citep{abbas2021computer} and autism spectrum disorders \citep{de2020computer}. Features derived from  overall facial expression, eyes, gait, and posture (Table~\ref{tab_feature})  have been found to be relevant across all of these disorders. 

Studies in suicidal ideation have mainly focused on using interpretable ML for characterizing the disorder. This makes the ML models more applicable in augmenting human caregivers by bringing up specific insight that they would like to measure. In depression studies, some approaches \citep{chen2021sequential} have also involved fusion  of video features derived from each frame that are used to train a sequential deep network and most have used pre-training to account for the relatively small size of depression datasets \citep{he2022deep}. While these models perform very well on the same held-out test set, their clinical applications remain limited due to a lack of interpretability. To improve interpretability, few researchers have built depression activation maps  to highlight the facial areas corresponding to depression severity as learned by the model \citep{zhou2018visually}. Meanwhile, using predefined features  has been most successful in providing interpretable results \citep{smrke2021language,he2022deep}.

\begin{table*}
\centering
    \caption{ Multimodal data features and their uses in mental health.}
  \scalebox{0.8}{
  \begin{tabular}{llll}
  \hline
	\bf Features &	\bf Example(s) &	\bf Example relevant in mental disorder(s)	     \\
  \hline
  
 \multicolumn{3}{c}{ Acoustic  } \\ \hline 
 
 Source	of sound features  	& Jitter 	&  Increase with depression severity \\
 
Filtering features by vocal and nasal tracks &	First resonant peak in the spectrum 	&  increase with bipolar severity\\

Spectral features  of speech &	Mel frequency cepstral coefficients & A variety of disorders \\

Prosodic features of speech &	Pause duration	& Higher in SCZ \\  \hline

  \multicolumn{3}{c}{ Video } \\ \hline 
 
Facial	& Smile duration, eyebrow movement, disgust expression &	Increased disgust expression in SI \\

Eyes &	Gaze angle &	More non-mutual gazes in MDD\\

Gait &	Arm swing and stride &	Reduced arm swing in MDD \\ 

Posture	& Head pitch variance, upper body movements	& Reduced head movement in SCZ\\

& & Higher head movement in ASD\\ \hline 

 \multicolumn{3}{c}{ Language  } \\ \hline 

Grandiosity	& Unrealistic sense of superiority	& Increased in bipolar \\

Semantic coherence	& Flow of meaning 	& Decreased in psychosis \\

Rumination 	& Repetitive thought patterns	 & Increased in MDD \\

Self-focus	& Self-referent information &	Increased in stress \\ \hline
\end{tabular}}
\label{tab_feature}
\end{table*}

\subsection{Natural Language Processing (NLP)}
NLP techniques enable computers to analyze, understand, and derive meaning from text and speech in a similar manner to humans. With NLP, mental health professionals can evaluate patterns in language to help identify and predict psychiatric illness in patients (Table~\ref{tab_feature}). Language is not only one of the primary expressions of human behavior that carries a variety of implicit and explicit markers  relevant to mental health \citep{rezaii2022natural,le2021machine}, but also more abundantly available compared to speech data. For example, social media platforms contain a large quantity of real-world language data, whereas the matched scale of speech data is limited. There are two sorts of NLP applications for detecting  specific mental health symptoms. The first type of applications is directly applied to patients, varying from predicting  the risk of suicide and early psychiatric readmission to identifying phenotypes and comorbidities. The second type of applications is indirectly applied to EHR and clinical records (tests, transcripts), which can be used for automating chart reviews, clustering patients into phenotype subtypes, predicting patient-specific outcomes. EHRs (including pathology reports, lab results, clinical tests and clinical session transcripts) are systematic collections of longitudinal, patient-centered clinical records. Patients' EHRs consist of both structured and unstructured data: the structured data include information about a patient's diagnosis, medications, and laboratory test results, and the unstructured data include information in clinical notes.

Massive EHR datasets have provided opportunities to adapt computational approaches to track and identify target areas for quality improvement in mental health care. According to a 2015 national survey, 61.3\% of US psychiatrists use EHRs \citep{cdc2015}. The EHR language is at least one level abstracted from the patient's symptoms, consisting of clinical notes by physicians. However, the unique advantage of EHR data is the ease with which demographic and socioeconomic features can be combined with language data. Symptoms derived from the free text in EHR data can be used to predict a diagnosis for  bipolar disorder  \citep{castro2015validation}, situational aggression \citep{van2018risk}, and  suicidal ideation \citep{walsh2018predicting},   with reported performances comparable to clinicians. Furthermore, discharge summaries from EHRs have also been used to predict remission \citep{rumshisky2016predicting}.  Aside from symptoms, a variety of relevant mental health data such as intervention status and physical health comorbidities can be routinely extracted from EHRs using NLP methods  \citep{stewart2021applied}. Privacy concerns around  EHR data sharing remain one of the key challenges in validating generalization of NLP methods. Encouragingly, there has been growing interests in using transformers for generating artificial mental health clinical notes to mitigate this issue \citep{stewart2021applied,ive2020generation}.

The advances in text-based mental health interventions (e.g., Talkspace and CrisisTextLine) have made transcripts of clinical sessions easily amenable via NLP. Aside from developing models for detecting suicide ideation  \citep{bantilan2021just}, NLP can also be applied to these datasets to understand the population-level trend, such as the increase in anxiety and decrease in quality of personal relationships during the COVID-19 pandemic \citep{raveau2022natural}.
Since language data are ubiquitous, one of the challenges in applying NLP for advancing mental health is data standardization. Depending on the task, different types of data may yield different levels of “signal”. For example, to predict the first episode psychosis, language data from clinical tests has higher performance  compared to transcripts of free speech \citep{morgan2021natural}. On the other hand, data collected “free-speech” samples for diagnostic purposes has been found to be highly effective in developing a language-based depression screening  that generalizes well across various age groups \citep{lu2020robust,rutowski2020depression,rutowski2021cross}.

\subsection{Social Media}
To date, social media companies have collected a wide variety and a large amount of language data which may contain clinically-relevant information (Table~\ref{tab_socialmedia}). This information can not only  be extracted on a population level, such as the notable rise in cognitive distortions over time \citep{bollen2021historical}, but also  be attributed on an individual level  \citep{bathina2021individuals}, making social media a powerful tool to support mental health risk assessment and diagnosis. Language from Facebook posts, for example, has been shown to contain markers for depression;  rumination and sadness can be detected in such data up to 6 months prior to a clinical diagnosis at hospital \citep{eichstaedt2018facebook}. Models applied to Facebook and other social platforms (e.g., Twitter and Reddit) have been successful in predicting diagnosis of  psychosis, anorexia, anxiety, and stress levels \citep{kim2020deep,guntuku2019understanding,rissola2021survey}. Aside from language present in the user posts and comments, ML models often process media data such as Instagram images \citep{hansel2021utilizing}, or integrate images and text to infer the user's  state-of-the-mind  \citep{birnbaum2020identifying,hansel2021utilizing}. Entries of online search also form a complementary and equally compelling dataset alongside social media activity.

Recent developments of transformer models, including those learning multilingual language representations, have enabled researchers to build NLP models generalizable across languages and apply them to social media data, for example, to detect depression or self-harm \citep{CairoDep2021,BERT2021}. Furthermore, specialized language representations that were trained on mental health specific conversations and became  publicly available \citep{ji2021mentalbert}, have been shown to improve performance compared to non-specific representations. Finally,  transformer embedding can be applied to pre-identified
sections in language which correspond to responses to standard clinical assessments such as the
subjective well-being scales \citep{kjell2022natural}, supporting the high-accuracy prediction of standard survey
scale responses  without directly running the survey.

While social media solves the scale issue with millions of samples available, most social media data lack  clinically-valid labels \citep{chancellor2020methods}. Most work has relied on using self-disclosure of mental illness as the labels, which tend to be noisy and may bring the additional issue of defining a healthy control. Despite the challenges, the validity of social media data has been repeatedly proven to support mental health diagnosis and risk assessment.

\begin{table}
    \caption{Social media data access (OAuth 2.0 is the industry-standard protocol for authorization).} 
    \resizebox{\columnwidth}{!}{
  \begin{tabular}{lll}
  \hline

\bf Platform  &	\bf Examples of accessible data  &	\bf OAuth 2.0 required \\ \hline
 
Instagram &	Media, captions, total posts &	Yes \\

TikTok	& Videos, descriptions, likes	& Yes\\

SnapChat& 	Bitmoji, public stories/media & 	Yes\\

Twitter	& Tweets, timestamps, likes, retweets	&  No\\

Pinterest& 	Pins, boards& 	Yes\\

Kik	& Messages through a bot& 	Yes\\

Tumblr& 	Blogs, profile, likes& 	Yes\\

Reddit	& Posts, post timing& 	No\\

Discord	& Conversations on a server& 	Yes\\

Facebook& 	Posts, media& 	Yes\\

YouTube	& Watch history, comments& 	Yes\\

Twitch& 	Chat history & 	Yes\\

Google & 	Query Strings& 	Yes\\

  \hline
\end{tabular}
}
\label{tab_socialmedia}
\end{table}

\subsection{Sensing Technologies and Mobile Mental Health}
Smartphones, wearables and other connected devices equipped with ambient sensors (Figure~\ref{fig_tech}B) are increasingly capable of recording physiological measurements  that are known to affect mental health \citep{garcia2018mental}. In addition, some of the less obvious measurements (such as keystroke usage patterns) have been shown to be implicated by mental illness \citep{dagum2018digital,zulueta2018predicting}. Additionally, online gaming behaviors, such as interaction patterns with non-player characters (NPCs) and other game behavior patterns, can be used to measure cognitive performance and their relationship with mental illness \citep{mandryk2019potential,Dechant2021}. 

\begin{table}
    \caption{Mobile sensor measurements and potential applications for mental health monitoring.} 
\resizebox{\columnwidth}{!}{
  \begin{tabular}{lll}
  \hline

\bf Measurement &	\bf Feature  &	\bf Effect in mental health \\ \hline
Movement&	Psychomotor agitation&	Increased in Anxiety \\
Location &	Social avoidance&	Increased in MDD  \\
Social activity	& Call/text volume &	Reduced in MDD \\
Keystroke  &	Keystroke latency & Impaired in ADHD \\
Heart rate &	Heart rate variability & 	Impaired in Stress \\
Gaming &	NPC interactions &	Impaired in social anxiety \\

  \hline
\end{tabular}
}
\label{tab_sensor}
\end{table}

These measurements from mobile sensors (Table~\ref{tab_sensor}) form valuable  sources of mental health data, and can be useful at various levels of granularity--from raw sensor data (e.g., the accelerometer) to derived high-level features (such as psychomotor activity). This has inspired many corporations to invent technologies for detecting depression and cognitive decline based on data collected from their wearable devices \citep{WSJ2021_apple}. Sensor-based measurements are found to be correlated with high-stress levels and a variety of ailments including depression, anxiety, psychosis, and bipolar disorder  \citep{seppala2019mobile,chikersal2021detecting}. Since sensor-based data are widespread and readily available, they offer an opportunity to build baseline models for individual users; these baseline models can be used to identify significant physiological changes in users and further inform clinical interventions.
       
Digital phenotyping for  individuals is  based on data acquired from mobile device collected in real time.  Devices that collect data streams from patients, such as surveys, cognitive tests, social medial interactions, GPS coordinates, and behavioral patterns (e.g., keyboard typing), have great potentials for monitoring, managing and predicting the individual’s mental health \citep{torous2016new,mohr2017personal}. Overall, continuous quantification of these data streams may result in clinically useful markers that can be used to refine diagnostic processes, tailor treatment choices, improve condition monitoring for actionable outcomes (such as early signs of relapse), and develop new intervention models \citep{huckvale2019toward}.

\begin{table}
    \caption{Commercial and research platforms and services for mental health applications.} 
    \resizebox{\columnwidth}{!}{
  \begin{tabular}{lll}
  \hline

\bf Platform  &	\bf Primary data source &	\bf Mental health appl. \\ \hline
 
 WoeBot \citep{prochaska2021therapeutic} &Language& 	Depression, Anxiety \\ 

Mindstrong \citep{dagum2018digital}	& Keystrokes &	Serious mental illness\\

Sonde Health \citep{huang2019investigation} &	Voice	&Mental fitness\\

Ellipsis Health \citep{harati2021speech}&	Voice&	Stress\\

Amazon Halo \citep{HaloVoice}	&Voice&	Emotion detection\\

Apple Watch \citep{Sonde} &	Mobility, Sleep&	Depression\\

Alphabet Fitbit \citep{Fitbit} &	Skin conductance&	Stress\\

Kintsugi \citep{Kintsugi} &	Voice&	Depression, Anxiety\\

Bewie \citep{torous2016new}
& Raw data from smartphone &	Multiple\\

MindDoc	& Language (ask daily question)&	Depression \\

Clarigent Health &	Voice	& Suicide risk \\

  \hline
\end{tabular}
}
\label{tab_platform}
\end{table}

\subsection{Commercial and Research Platforms and Services}
While  studies have demonstrated promising results in using ML to support the patient's journey in mental health, the applicability in clinical practice remains limited. Table~\ref{tab_platform} lists examples of platforms and services that use ML for mental health. While most platforms focus on developing risk assessment based on  single modality, the initial commercial viability of these platforms is promising for the success of using ML in mental health since they enable collection of large amounts of data which can be used in further development of biomarkers.

Many areas of mental health technology development focus on scalablity in  clinical research. For example, there are between 10,000 to 20,000 smartphone apps that digitize mindfulness or cognitive behavioral therapy techniques \citep{DeloitteAppMarket}, allowing the  user to engage in psychotherapy on their own at a greatly reduced price compared to in-person therapy. However, the quality is highly variable and the mechanisms used to validate them is often dubious. Moreover, since this area is relatively new, the industry and governmental standards to validate such a technology are still in the early phase. We will briefly outline two interrelated areas of development: digital measurements and digital interventions.

\textbf{Digital measurement applications.} We are entering a new era of digital psychiatry \citep{burr2020digital,torous2021digital}. In 2016, the Harvard professor Jukka-Pekka  Onnela coined the term {\it digital phenotype} \citep{onnela2016harnessing}, which refers to the use of mobile devices and other digital data sources to measure behavior and physiology for understanding brain activity that is relevant to pathological states. These techniques utilize measurement paradigms from translational neuroscience that were developed in laboratory settings such as direct quantification of motor (i.e. movement, muscle activation) and physiological activity (i.e. heart rate, electro-dermal response) more than traditional clinical scales or self-report scales. The advantage of this approach is that it better aligns with emerging knowledge of rapid-acting biological processes and provides high measurement accuracy through direct rapid sampling, which is in  contrast to  traditional clinical measures that are taken sporadically  over a long  period \citep{stein2021mental,Abbas2021,galatzer2013636}. These measurement approaches have relevance in multiple areas  including treatment development, treatment selection, and ongoing monitoring.

Medications that target mental health conditions have a significant history of failure. Most psychiatric medications were discovered capriciously rather than being developed based on knowledge of the underlying biological mechanisms. As new medications emerge from basic and translational neuroscience research, both drug developers and clinicians struggle with how to measure the effects of new treatments and how to properly target old treatments. For example, traditional antidepressant medications are designed to slowly titrate serotonin levels, resulting in slow global effects over a 2-4 week period. Correspondingly, measures of depression based on the DSM, query about the presence of depressive states over a 2-week period. New classes of anti-depressant such as ketamine and psilocybinpsylocibin affect specific depressive symptoms in minutes. Further, the mechanistic effects, and thus the need for measurement, is much more specific and granular. In fact, most classes of anti-depressants including serotonin reuptake inhibitors (SSRI) and psilocybinpsylocibin, and  ketamine, act on serotonin receptors that ultimately impact peripheral motor and physiological activity \citep{jacobs1994serotonin,gigliucci2013ketamine}. Serotonin regulation will likely have a direct effect on depression symptoms such as psychomotor retardation, but the  direct effect on feelings of guilt is minimal. As such, methods used to directly measure motor output have a higher likelihood of capturing both pathology and treatment effects.

As an example, research effort has been dedicated to using computer vision and voice to directly quantify motor activity. Some recent work has demonstrated that digital phenotyping parameters that reflect gross motor activity including speech characteristics (rate of speech, tone) and facial/head movements are associated with suicidal risk \citep{Issac2021}, SSRI response in MDD \citep{abbas2021remote}, negative symptomatology in schizophrenia \citep{Abbas2022}, and Parkinsonian tremor \citep{zhang2021estimation}. Such approaches are now being commercialized for all phases of drug development from proof-of-concept to direct measurement to make decisions about ongoing treatment needs. Such measures solve many of the current problems in clinical measurement since they  can be captured remotely in an automated way. These measures can also be captured at a much higher frequency and can provide a sensitive numeric value.

These new approaches to measurement have significant challenges. First, methods that are adapted from the laboratory often lack the tight experimental control necessary to interpret the data correctly. For example, a rapid increase in physiology can indicate stress, but also exercise or other forms of exertion. Second, while the scientific basis of these measurement paradigms may be sound, commercial approaches are rarely validated to the extent required to be of clinical utility and rarely sufficiently transparent in their approach to be used for regulatory approval.  

\textbf{Digital interventions.} The other rapidly emerging area of mental health technology are digital approaches to clinical care. We will briefly outline some of the leading approaches. Importantly, digital approaches to clinical care are often aligned with a digital measurement approach as these approaches are ''blind'' without some sort of remote data. A number of companies such as Mindstrong Health \citep{dagum2018digital}, IesoTrigger Health \citep{ewbank2020quantifying}, and Headspace Health \citep{economides2018improvements,kunkle2021association}, have attempted to integrate digital phenotyping  to identify when patients are in acute clinical need. However, it is unclear how accurate these methods are as they are typically unpublished. This has led to development of models that can identify patterns in patient's and clinician's  language   that are markers of improved outcomes \citep{ewbank2021understanding}, which can be further used to measure success of various therapy modalities, treatment design, as well as to improve care quality \citep{flemotomos2021automated}.

Digital therapeutics proposes the use of mobile devices to offer automated cognitive behavioral therapy (CBT), mindfulness, or other validated psychotherapy  in an automated fashion. These app-based approaches undergo the same clinical validation process and traditional medications, and are often developed in collaboration with large drug developers. Examples that are in development or have received FDA (Food and Drug Administration) approval include treatments for substance use disorder (SUD), ADHD, schizophrenia, ASD, MDD, PTSD, and generalized anxiety disorder (GAD) \citep{patel2020characteristics}.

While digital therapeutics attempt to scale treatment, telehealth aims to scale the treatment provider network. Mental health treatment is an area where there are many effective treatments but little access to treatment providers \citep{insel2019bending}. The issue of access became most acute during the emergence of COVID-19 when significant wide-spread mental health needs emerged along with greatly decreased access to care. To address this need, a large array of options have emerged, many reinforced by the emergency COVID-19 Telehealth Act of 2021 that enabled remote patient care \citep{folk2022transition}. These services, which are accessible directly to consumers, or more often, provided by a third-party payer, provide access to coaches and clinicians via different mobile platforms including text, voice, and video communication. Services vary from mental health coaching provided by lay-professionals to psychiatric and psychological services. While still in their infancy, early evidence has shown that telehealth services can perform at parity with traditional in person therapy \citep{wagner2014internet}. Therefore, teletherapy is likely to be a dominant form of mental health treatment in the future.

\section{Multimodal Data fusion in Diagnostic Analytics}
A central goal of precision psychiatry is to integrate all clinical, physiological, neuroimaging, and behavioral data to derive reliable individualized diagnosis and therapeutics. Importantly, the health-related data are produced daily, especially from personal devices. The most essential effort in multimodal data analysis tasks is to explore the relationship between modalities, complementarity, shared versus modality-specific information and other mutual properties. Multimodal data fusion techniques present a framework to infer information how different data modalities interact and can be integrated for improved disease prediction \citep{lahat2015multimodal,croitor2011fusing,calhoun2016multimodal}. In this section, we will review several data fusion methods in diagnostic analytics (Section \ref{section-Multimodal-1}). We will focus on multimodal neuroimaging data (Section \ref{section-Multimodal-2}), and then extend the discussion to other modalities including vocal and visual expression data (Section \ref{section-Multimodal-3}).

\subsection{Popular ML Methods for Multimodal Fusion}
\label{section-Multimodal-1}
In the past decades, numerous research efforts have been dedicated to developing powerful ML methods for multimodal data fusion \citep{adali2015multimodalICAIVA,adali2015multimodal,calhoun2016multimodal,calhoun2020multimodal,zhang2020advances}. Some commonly used approaches are summarized below.

\textbf{Multivariate Correlation Analysis.} Canonical correlation analysis (CCA) is a standard statistical method based on second-order statistics for data fusion. It aims at finding a pair of linear transformations to drive latent variables (aka. canonical variates) that have maximized correlation between two different data modalities \citep{correa2010canonical}. For a more general setting, multiset/multiway CCA (mCCA) has been developed as an extension of the standard CCA to multimodal fusion by maximizing the overall correlation among latent variables from more than two sets of modalities \citep{de2019multiway,calhoun2016multimodal}. Similar to CCA, partial least squares (PLS) and its extensions, i.e., multiway PLS (N-PLS), provide alternative approaches to integrate multimodal data by maximizing the covariance between latent variables from different modalities \citep{chen2016joint,silva2019integrate}.

\textbf{Matrix and Tensor Factorization.} Based on matrix and tensor factorization techniques, joint blind source separation (BSS) approaches have been developed and successfully applied to multimodal fusion of biomedical data \citep{zhou2016linked,adali2015multimodal}. As a typical example, joint independent component analysis (jICA) aims to maximize the independence among jointly estimated components from multiple modalities that are assumed to share the same mixing matrix \citep{calhoun2008feature}. The jICA approach involves concatenating modality features alongside each other and then performing ICA on the composite feature matrix \citep{calhoun2009review}. Independent vector analysis (IVA) is another extension of ICA to multiple datasets. IVA makes use of dependence across datasets by defining source component vectors concatenating a specific source estimated from multiple modalities \citep{adali2015multimodal,adali2015multimodalICAIVA}. Coupled matrix and tensor factorization (CMTF) was also developed to simultaneously factorize multiple datasets in the form of matrices and high-orders tensors using tensor decomposition \citep{acar2015data}, showing strength in capturing the potential multilinear structure for multimodal fusion. Besides extracting shared common components, some multimodal fusion tasks are also interested in deriving individual components that are modality-specific. Common and individual feature analysis (CIFA) \citep{zhou2015group} and joint and individual variation explained (JIVE) \citep{lock2013joint} models have been proposed to achieve this goal. By jointly decomposing multiple feature matrices, CIFA and JIVE are able to simultaneously estimate common and individual feature subspaces. A further extension of CIFA has been achieved by leveraging high-order tensor factorization \citep{zhou2016linked}, which provides an efficient way to perform a multidimensional fusion of multiple data modalities. 

\textbf{Multi-Kernel Learning.} Multi-kernel learning (MKL) has won many successful applications in multimodal data fusion due to the full utilization of multiple kernels that enable simultaneous learning from various modalities with heterogeneous data \citep{rakotomamonjy2008simplemkl,mariette2018unsupervised}. Different kernels naturally correspond to different modalities, such as neuroimaging, clinical, behavior, speech features, etc., which may provide complementary information to drive improved modal learning performance. The MKL problem can be set as a linear combination of kernel matrices or a nonlinear function with specified forms of regularization. MKL may be designed under different ML models including SVM, Gaussian process, and clustering. Among them, MKL-SVM has been most popularly applied to integrate heterogeneous data modalities in studies of mental health \citep{squarcina2017classification,dyrba2015multimodal,zhang2011multimodal}.

\textbf{Deep Learning-based Fusion.} Empowered by cutting-edge deep learning techniques, emerging methods have been increasingly developed for deep multimodal fusion \citep{ramachandram2017deep}. Data fusion through deep learning allows integrating multiple modalities based on learned high-level feature representations that are theoretically more comparable to each other and more informative for predicting the targets \citep{zhou2019effective}. By exploiting cross-modal manifolds as a feature graph, a deep manifold-regularized learning model was recently designed to integrate transcriptomics and electrophysiology data from neuronal cells, and yield promising performance for phenotype prediction \citep{nguyen2022deep}. Graph neural networks show capability in information fusion for multimodal causability by defining casual links between features with graph structures, thereby enhancing the explainability of the derived multimodal feature representation \citep{holzinger2021towards}. By extending graph neural networks to multimodal structures, deep representation approaches have also been designed for integrating brain networks constructed from diverse modalities \citep{dsouza2021m,zhang2020deeprep,kong2021multiplex}.

\begin{figure}[!t]
\centering
\includegraphics[width=0.48\textwidth]{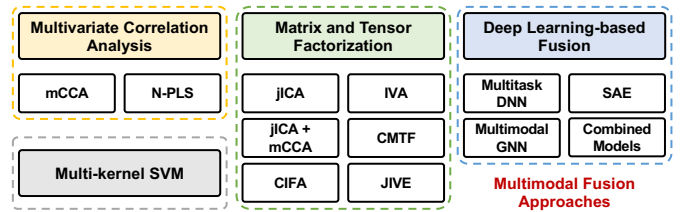}
\caption{Summary of typical approaches for multimodal data fusion in psychiatry studies.}
\label{fig-MultimodalMethods}
\end{figure}

\subsection{Multimodal Neuroimaging Studies}
\label{section-Multimodal-2}
Neuroimaging data types are intrinsically dissimilar in nature, having different spatial and temporal resolutions \citep{calhoun2020multimodal}. Instead of entering the entire data sets into a combined analysis, an alternate approach is to reduce each modality to low-dimensional (latent) features of selected brain activity or structure and then explore associations across these feature sets through variations across individuals. Exploiting such latent feature representations from multiple neuroimaging modalities for diagnosis has generally shown to improve performance compared to using a single modality alone \citep{tulay2019multimodal}. Multimodal fusion allows for the integration of neuroimaging data modalities from different scales of spatial and temporal resolutions. Combining multimodal neuroimaging offers an elegant way to exploit complementary information for more accurate and robust characterization of brain dysfunctions, and hence is instrumental in optimal decisions for diagnosis, treatment, prognosis, and planning in many applications in medicine.
% Therefore, multimodal learning mechanisms that take advantage of multimodal data sources are instrumental in optimal decisions for diagnosis, treatment, prognosis, and planning in many applications in medicine.

A combination of mCCA and jICA was successfully applied to fMRI and DTI fusion in the diagnosis of schizophrenia and bipolar disorder \citep{sui2011discriminating}. CMTF has been applied to identify diagnostic biomarkers of schizophrenia by integrating sMRI, fMRI, and EEG \citep{acar2019unraveling}. MKL-SVM has been successfully applied to integrate multimodal structural neuroimaging for predicting differential diagnosis between bipolar and unipolar depression \citep{vai2020predicting} and to combine sMRI and fMRI for improved classification of trauma survivors with and without PTSD \citep{zhang2016multimodal}. More recently, it also showed efficacy in the diagnosis of early adolescent ADHD by integrating sMRI, fMRI, and DTI \citep{zhou2021multimodal}. In learning low dimensional representations of functional and structural MRI \citep{geenjaar2021fusing}, the functional MRI can be split into several independent component networks, each treated as a separate modality along with the structural scan for learning using autoencoders. Furthermore, MKL methods have been used for diagnosing schizophrenia by combining markers from MRI and DTI \citep{liu2018mmm}. A multimodal graph convolutional network was designed to integrate functional and structural connectomics data for an improved prediction of phenotypic characterizations in ASD \citep{dsouza2021m}. By combining multiple typical neural network structures, multimodal deep learning models have also been developed to effectively integrate fMRI connectivity and sMRI features \citep{plis2018reading}, and also genomic data \citep{rahaman2021multi} for discovering schizophrenia-associated brain dysfunction. Methods for learning joint representations from neuroimaging and non-neuroimaging data are still in early development \citep{akhonda2022association} and there is an opportunity for ML methods to evolve for this task. For example, transformer networks with late fusion can be used to learn joint representations from various modalities such as EEG and eye movement signals \citep{wang2021emotion}.

\subsection{Multimodal Fusion of Non-imaging Data}
\label{section-Multimodal-3}
Multimodal approaches consist of combining data from various sources to jointly arrive at an answer. Given how little is conclusively known about which type of data, neuroimaging, social media, speech, video, sensor data carries the most phenotypes for mental illness, it only makes sense to combine the information from these data sources. In addition, this also enables modeling the inter-dependencies between these data which may not be observable by a human expert at the same time. For example, many features listed in Table~\ref{tab_feature} are known to be relevant in depression; however, observing them simultaneously can be very challenging for a clinician. MKL and multi-task learning methods have been used to jointly learn from sensor and smartphone usage data to predict subjective well-being \citep{jaques2015multi}. The success of transformer networks in jointly modeling video, speech, and language data has catalyzed multimodal modeling in mental health \citep{lam2019context}. Multimodal modeling techniques can also be used in modeling symptoms such as emotion dysregulation \citep{parra2022}, loneliness \citep{doryab2019identifying} and sentiment analysis \citep{he2021unimodal}. In a prognostic study, an SVM-based multimodal ML approach was developed to integrate clinical, neurocognitive, neuroimaging, and genetic information to predict psychosis in patients with clinical high-risk states \citep{koutsouleris2021multimodal}. Deep autoencoder-based fusion approaches have been designed to integrate dynamics of facial and head movement and vocalization, and successfully applied to the prediction of depression severity \citep{dibekliouglu2017dynamic}.

\section{ML for Molecular Phenotyping in Psychiatry}

\begin{figure}[!t]
\centering
\includegraphics[width=8.5cm]{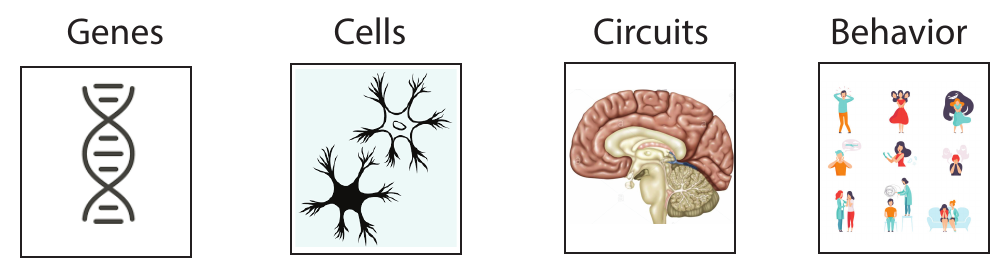}
\includegraphics[width=9.5cm]{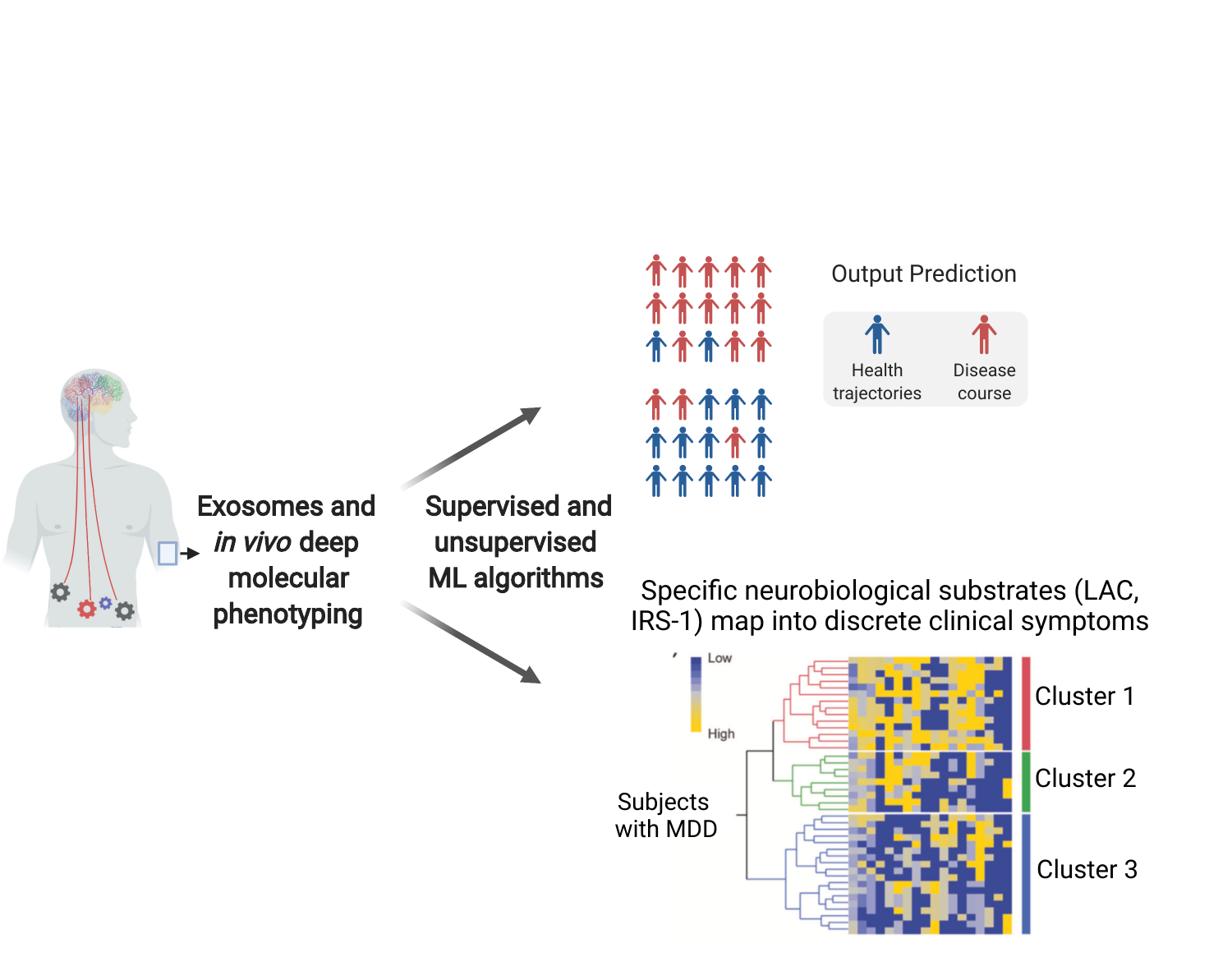}
\vskip -5mm
\caption{(A) Different levels of interacting variables (genes, cells, circuits) to behaviors in mental illnesses. 
(B) Combining ML with novel molecular biology technologies (for deep molecular phenotyping of brain plasticity) opens up  opportunities to develop new mechanistic models for prevention and treatment of clinical endophenotypes of mood and cognitive disorders.}
\label{fig_molecular}
\end{figure}

Molecular phenotyping is referred to as the technique of quantifying pathway reporter genes (i.e. pre-selected genes that are modulated specifically by metabolic and signaling pathways) in order to infer activity of these pathways. Mapping genes and genomics to behaviors can identify risk factors and biomarkers in mental disorders. 
The brain is the central organ exposed to stressors and external behavioral interventions, and therefore is a vulnerable organ subject to changes in multiple interacting biological networks at the systems level. ML methods have been contributing enormously to capturing the complexities of interacting variables within and across multiple levels (Figure~\ref{fig_molecular}A)-- particularly at the molecular level -- for identifying mechanistic-based phenotyping models as new targets for the prevention and treatment of mood and cognitive disorders. The advent of unbiased next-generation sequencing (NGS) prompted the development of advanced bioinformatics and ML  tools to profile and decode large molecular datasets (e.g.: transcriptomics, epigenomics, metabolomics) at the genome-wide level in health and disease states (Figure~\ref{fig_molecular}B). To date, there is increasing recognition of the utility of ML methods to integrate these multi-level molecular datasets with clinical characteristics to map specific neurobiological substrates into the complexity of symptom clusters, which may aid the classification of diseases, prediction of treatment outcomes, and selection of personalized treatment. Emerging avenues in molecular neuroscience and computational psychiatry for new mechanistic models for the prevention and treatment of CNS disorders.

Animal models has been playing a vital role towards precision psychiatry in understanding mechanisms of diseases and predicting treatment responses \citep{herzog2018,bale2019}. To bridge the scientific knowledge gap from animals to humans, gene expression studies offer an example of how integrating basic neuroscience, ML, and bioinformatics approaches can contribute to advancing understanding of the molecular basis of MDD. Using RNA sequencing (RNA-seq) assays and gene coexpression network analyses (based on hierarchical clustering to identify gene modules), differential gene expression profiles have been shown across six key brain regions, such as the ventromedial prefrontal cortex (PFC) and dorsolateral PFC among others, in post-mortem tissues of subjects suffering from MDD as compared to age and sex-matched controls with remarkable sex differences in these molecular pathways \citep{RN1}. Recent work using RNA-seq assays at single nucleus resolution (snRNA-seq) and t-distributed stochastic neighbor (t-SNE) embedding analyses showed cell-type specific transcriptomic profiles in the post-mortem dorsolateral PFC that are differentially regulated in MDD cases as compared to respective controls, with the greatest gene expression changes in deep layer excitatory neurons and immature oligodendrocyte precursor cells \citep{RN2}. Importantly, these gene expression studies in humans were supported by findings in rodents showing a brain that continually changes with experience \citep{RN3}. Several groups using RNA-seq assays and bioinformatic analyses showed striking transcriptomic differences in the ventral and dorsal hippocampus in the responses to stress--a primary risk factor for multiple psychiatric diseases--with the ventral hippocampus being particularly sensitive not only to the effects of stress \citep{RN4} but also a target for the responses to next-generation antidepressants \citep{RN5,RN6}.

The expansion of NGS to single-cell resolution assays requires more advanced bioinformatics analyses, which utilize  ML  to analyze these large datasets, including denoising and dimensionality reduction,  cell-type classification, gene regulatory network inference, and multimodal data integration \citep{RN7,RN8}. Bioinformatic approaches, such as Seurat, combine unsupervised nonlinear dimensionality reduction, K-nearest neighbor graph analysis for cell-type clustering, and weighted nearest neighbor analysis for multimodal data integration \citep{RN9}. Deep learning approaches, such as the autoencoder, provide another example of tools used for denoising and dimensionality reduction with computational scalability \citep{RN10,RN11,RN12}. Autoencoders can also be used in a supervised manner for transfer learning across datasets, e.g., to learn the embedding from a larger, already annotated dataset and transfer this knowledge to cluster new datasets \citep{RN13}. The combination of multimodal data generated from the simultaneous assessment of transcriptomic profiles with regulatory landscape or spatial location in the same single cell \citep{RN14,RN15,RN16,RN17} will allow a deeper molecular characterization of discrete cellular states \citep{RN9}.

\textbf{Integrating multidimensional factors for new mechanistic treatment models.} It has been increasingly recognized that mood and cognitive disorders are unlikely to be only brain-based diseases. Additionally, growing evidence suggest that they are system-level disorders affecting multiple interacting biological pathways \citep{RN18}, involving the dynamic cross-talk between the brain and the body. Using hierarchical clustering to integrate {\it in-vivo} molecular measures of brain metabolism with clinical symptoms in MDD patients, recent work showed that, at least in these cases, the specific neurobiological substrates map into discrete clinical symptoms, including anhedonia \citep{RN19}. Furthermore, the integration of multidimensional factors spanning mitochondrial metabolism, cellular aging, metabolic function, and childhood trauma has been shown to provide more detailed signatures to predict longitudinal changes in depression severity in response to the metabolic agents used as antidepressant treatment than individual factors \citep{RN20}. Use of multi-omics approaches and random forest classifier has  been shown to achieve  85\% sensitivity and 77\% specificity in prediction of the PTSD status. This system-level diagnostic panel of multiple molecular and physiological measures outperformed separate panels composed of each individual data type, with certain mitochondrial metabolites among the most important predictors \citep{RN21,RN22}.

An additional ML application example  include the integration of multidimensional phenotypic measures to identify those mechanisms that predispose apparently healthy individuals to develop maladaptive coping strategies from those that confer resilience. Using a high-throughput unbiased automated phenotyping platform that collects more than 2000 behavioral features and supervised ML that minimizes Bayesian misclassification probability, recent work showed that a rich set of behavioral alterations distinguish susceptible versus resilient phenotypes after exposure to social defeat stress (SDS) in rodents \citep{RN23,RN24}. At the individual level, a ML classifier integrated {\it a priori} constructs, including measures of anxiety and immune system function, predicted if a given animal developed SDS-induced social withdrawal, or remained resilient, with 80\% sensitivity, which  is better than the categorization power based on either individual measure alone \citep{RN25}. 

To develop personalized psychiatry strategies for better diagnosing and treating mental disorders, it is essential to meet the demand for ML  enforced by the recent advent of molecular biology protocols that opened up the opportunity to capture central nervous system nanovesicles (known as {\it exosomes}) for examining specific neurobiological substrates, including transcriptomic profiles, dynamically and temporally {\it in vivo}. The development of advanced ML methods for dynamic network analyses will permit to link brain molecular targets and signaling pathways with other levels of analyses (e.g., functional connectivity) {\it in vivo} and to incorporate complex relationships between the brain and the rest of the body to redefine thinking about the modifiable mechanisms throughout the complex clinical disease course.

\section{Explainable AI and Causality Testing in Psychiatry}
\label{section-XAI}
Explainable Artificial Intelligence (XAI) aims to provide strong predictive values along with a mechanistic understanding of AI by combining ML techniques with effective explanatory techniques and make it easily understood by the end users. XAI has found emergent applications in medicine, finance, economy, security, and defense \citep{gunning2019darpa,roessner2021taming}. In psychiatry, the mission of XAI is to help clarify the link between neural circuits to behavior, and to improve our understanding of therapeutic strategies to enhance cognitive, affective, and social functions \citep{lancet2020_XAI,sheu2020illuminating}. XAI distinguishes the standard AI in two important ways: (i) promote transparency, interpretability, and generalizability; (ii) transform classical ``black box" ML models into ``glass box" models, while achieving comparable or improved performance. From the diagnosis or prognosis perspective, it is crucial to know whether the ML solutions can be explainable to the point of providing hidden mechanistic insights into the way brains execute a particular function or complex behaviors. For instance, a classification function learned by the machine to predict a disease outcome would not only need to report a probability outcome but also need to address  additional questions for the end-user: why is this outcome instead of the alternative? How reliable is the outcome? When does it fail if something is missing or misrepresented? When and why the prediction is wrong? Accordingly, a model with improved interpretability is often accompanied  with parameter/structure/connectivity constraints and some prior domain knowledge. The models can be continuously adapted such that an iterative process  may be required to force ML methods to fit models with specific interpretations. From the treatment perspective, an improved understanding of brain dynamics responsible for dysfunctional cognitive functions and/or maladaptive behaviors in mental illnesses is also critical. To find the hidden cause, it is useful to discuss the concept of “causality”. 

Neuroimaging  provides a passive sensing approach to observe the (correlational) brain-behavior relationship. However, correlation is different from causation. Correlational dependencies describe associations of measurements that experiments do not control, whereas causal dependencies link a dependent variable to an experimentally controlled variable. The key concept in causal inference is to introduce randomization to perturb the mapping. The relationship between every dependable variable and the randomized variable is causal, whereas the relationship between non-randomized variables and behavior, remains correlational \citep{jazayeri2017navigating}. Closed-loop experimental design would help to test the potential causality \citep{chen2021improving}. In human experiments, we classify closed-loop testing into two categories: one being fully automated, and the other being closed-human-in-the-loop. 

\begin{figure}[!t]
\centering
\includegraphics[width=9cm]{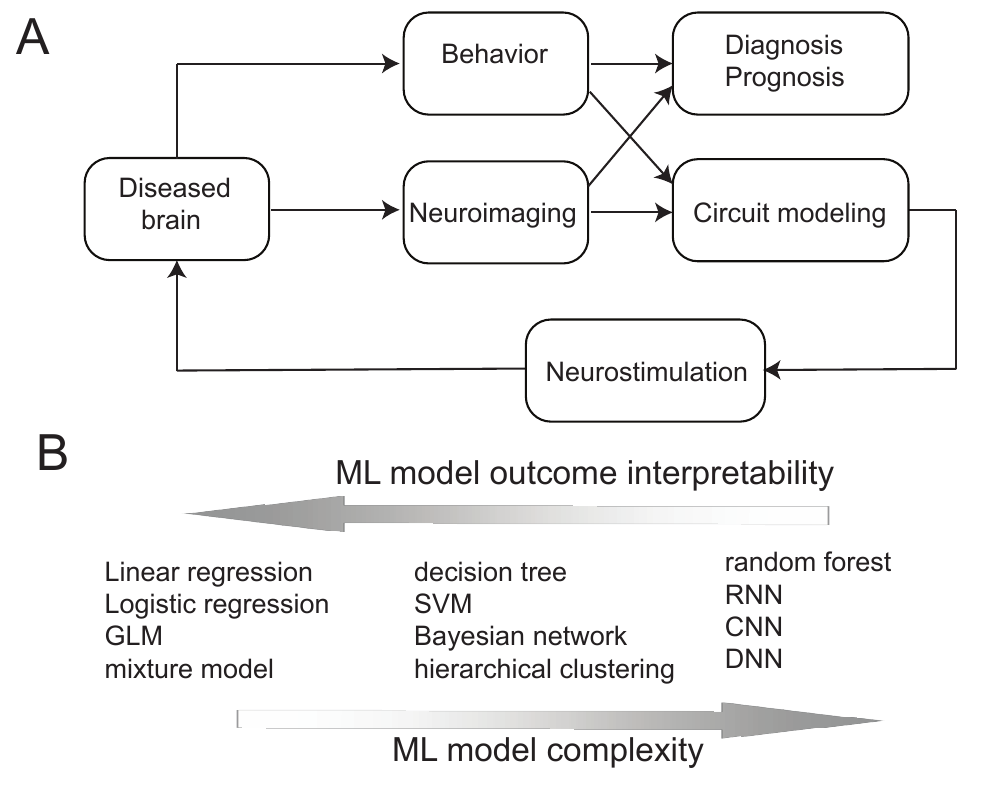}
\caption{(A) Schematic of the closed loop of neuroimaging/modeling/neurostimulation. (B) A wide spectrum of  interpretability in representative ML models. }
\label{fig_XAI}
\end{figure}

In what follows,  we will systematically review several important topics along these lines. We will present some ``glass box'' ML models in the literature (Section \ref{section-XAI-1}), then introduce  neuroimaging-based circuit-level modeling (Section \ref{section-XAI-2}) and adaptive neuromodulation (Section \ref{section-XAI-3}). We argue that combining these efforts in the loop (“Neuroimaging $\rightarrow$ Circuit modeling $\rightarrow$ Neurostimulation $\rightarrow$ Observing behaviors $\rightarrow$ Revising models”, Figure~\ref{fig_XAI}A) will provide an effective way towards interpretable neuropsychiatry in understanding brain-behavior causation.

\subsection{Explanation and Constraints in Interpretable ML Models}
\label{section-XAI-1}
In terms of taxonomy, intrinsic interpretability refers to ML models that are considered interpretable due to their simple structures, such as short decision trees or sparse linear models (Figure~\ref{fig_XAI}B). Post-hoc interpretability refers to the application of interpretation methods after model training \citep{molnar2022}. Interpretation may appear with different forms: (i) finite feature summary statistics, (ii) meaningful model parameters, (iii) ease of visualization of model outcome (e.g., feature summary or decision boundary). An ML model that is highly interpretable because it has a few or more of the following properties \citep{molnar2022}: high expressive power, low translucency, high portability, low algorithmic complexity and informative constraints. Generally, there is a trade-off between explainablity and performance. For instance, a constrained linear or bilinear model will fit many of these criteria, but the linear model does not warrant a good performance. Additionally, a model that is potentially explainable does not guarantee explainability. For example,  co-dependence of input variables may make explanations ambiguous; latent variables of probabilistic generative model may face the problem of “explaining away”  \citep{Pearl1988}. Here we briefly mention several classes  of interpretable ML models.

\textbf{Hybrid rule-based ML models:} This type of ML models can be used for generating rules, such as a decision rule set: \texttt{IF (condition) THEN (outcome 1) ELSE (outcome 2) statement}, where the conditional clause will be learned from the data \citep{Letham2015}. This type of model has more expressive power but less portability. 

\textbf{Constrained ML models:} This type of ML models imposes parameter constraints to either avoid overfitting or enhance interpretability. Examples of such include the constrained  convolutional filters in the CNN \citep{Li2017NIPS}, or constrained mixture models \citep{zou2012diversity}. As a result, these constrained models have low translucency. 

\textbf{Feedback ML models:} ML models can be provided with user feedback in the human-in-the-loop system, where the user feedback is treated as constraints in the optimization problem \citep{Nair2021IJCAI,Daly2021AAAI}. The feedback can appear as a form of rule sets, which can be either known or unknown in advance. Iterating the feedback-rule optimization can generate more accurate rule sets. This type of model has good expressive power and high portability.

%We then trained an ML algorithm and other statistical learning techniques to predict the risk of death. The ML algorithm predicted mortality with an area under receiver operating characteristic curve (AUROC) of 0.80. We used class-contrastive analysis to produce explanations for the model predictions. We outline the scenarios in which class-contrastive analysis is likely to be successful in producing explanations for model predictions \citep{Banerjee2021} 

\subsection{Circuit-Level Modeling for Computational Psychiatry}
\label{section-XAI-2}
Sharing the same goal of XAI, computational psychiatry tries to combine multiple levels and types of computation with multiple types of data in an effort to improve understanding, prediction and treatment of mental illness \citep{huys2016computational}. Two complementary approaches are used in computational psychiatry: (i) the data-driven approaches apply ML methods to high-dimensional data, multimodal to tackle classification and prediction problems (Section \ref{section-WhatHow}); (ii) the theory-driven approaches develop empirical or mechanistic models to test hypothesis.

For mechanistic modeling approaches, circuit-level modeling of macroscopic or mesoscopic brain dynamics represents an important research topic in computational models for psychiatric disorders \citep{breakspear2017dynamic,murray2018biophysical,Papadopulos20}. A common strategy is to first use a biologically-inspired model to simulate neural activity based on a network of interacting neural masses, and then within each brain area, to model the neuronal population activity as the Wilson-Cowan neural mass  model, with each consisting of excitatory and inhibitory populations \citep{wilson:biophysj72}. Furthermore, individual brain nodes are coupled according to the empirically-derived anatomical network \citep{chaudhuri2015large}. The computational model can be driven by hypothesis or EEG/fMRI data. 

One data-driven macroscopic level modeling approach is Dynamic causal modeling (DCM). DCM has been widely used in characterizing the effective connectivity of a functional network based on task or resting-state fMRI \citep{Friston03,Friston14}, where the model parameter estimation and inference is done by unsupervised learning. By incorporating prior knowledge or hypotheses of network connections, DCM may reveal important brain mechanisms and offer experimental predictions. One of potential applications of DCM is to characterize the neural plasticity in the human brain, especially the change in functional connectivity informed by neuroimaging studies. The functional connectivity can either change gradually during the course of tasks or behavioral protocols, or induced by perturbation or neurostimulation. These changes are often, but not always, associated with changes in functional activations of specific brain regions. Combination of neuromodulation and DCM may provide a way to test the impact of neurostimulation on neural plasticity that underlies the change in adaptive or maladaptive behaviors.

\subsection{Closing the Loop for Testing Causality}
\label{section-XAI-3}
One big challenge in human psychiatric neuroscience is the causality gap \citep{etkin2018addressing}. Statistical causality or Granger causality between two variables is not equivalent to brain-behavioral causality. To identify an effective treatment strategy for mental illnesses, it is critical to causally modulate neural circuitry that is responsible for maladaptive behaviors. Human neuroimaging alone only demonstrates correlations but not causation. To understand the causal mechanisms, it is important to close the loop in experiments by manipulating or perturbing the brain circuits and measuring its outcome, as commonly done in animal experiments \citep{jazayeri2017navigating,chen2021improving}. Unfortunately, a rigorous and causal grounding of clinical symptoms and behavior in specific neural circuit alternations is still missing. Specifically, since the clinical symptoms are diverse, how to define the dimension of brain function that defines one or few clinical symptoms and effectively manipulate them remains unknown.

Temporally precise neurostimulation tools provide a plausible means to perturb or stimulate the brain. To date, human neuromodulation methods include invasive deep brain stimulation (DBS), noninvasive transcranial magnetic stimulation (TMS), noninvasive transcranial direct/alternating current stimulation (tDCS/tACS), and transcranial focused ultrasound stimulation (tFUS). A review of advances in neuromodulation technologies for treating mental disorders can be found in the literature \citep{lewis2016brain,romei2016information,Lo17}. To date, repetitive TMS (rTMS) has been cleared by the FDA for the treatment of depression and recently used in the studies of neural functioning and behavior \citep{Chen13PNAS,hobot2021causal}. 

The brain connectivity and dynamics can be studied from a network communication and control perspective \citep{tang2018colloquium,srivastava2020models}. The distinction between a healthy and a pathological brain can be characterized by their different  efficiency to route the information between distributed brain nodes, control/modulate the target node under specific constraints, or influence its behavior in order to perform specific tasks ("cognitive control") \citep{zhang2020data}. Brain connectivity and neurostimulation can be studied by applying the network and graph theory. Specifically, the control-theoretic models have also been applied to quantify the response of brain networks to exogenous and endogenous perturbations. Within the XAI-neuromodulation framework, it is convenient to formulae a mathematical framework for important research questions: (i) can a target node stimulation rewire the brain connectivity in evoked and steady-state conditions? (ii) can the neurostimulation-induced change of evoked or resting-state brain connectivity distinguish a pathological from a healthy brain? (iii) Given an input constraint (such as the energy of neurostimulation), what is the optimal neurostimulation policy? Can alternate or simultaneous neurostimulations at multiple sites more effectively influence the network connectivity or bring additional benefit in treatment \citep{fox2014resting}?

XAI for neurostimulation in mental health can be seen as an extension in the design of human brain-machine interface (BMI) \citep{fellous2019explainable}. Specifically, XAI can search and identify behaviorally activated targets through active and scheduled stimulation strategies. Traditional neurostimulation strategies are designed in an on/off stimulation fashion triggered by predetermined stimulation parameters. However, these stimulation parameters will not be optimal. To accommodate an adaptive subject-specific stimulation strategy, adaptive stimulation uses neurofeedback to adjust the stimulation parameters or control policy for achieving various optimality criteria; {\it reinforcement learning} and {\it active learning} (two other important ML topics not reviewed here) can play a guiding role in online adaptive experiments \citep{pineau2009treating,Tafazoli_2020}. Additionally, simultaneous or post-stimulation neuroimaging  provides a window of examining the change in brain network connectivity patterns. Can the induced brain patterns or changes in network connectivity be used to predict the treatment outcome? ML may address such a question by providing individualized treatment-response likelihood in precision psychiatry \citep{zandvakili2019use}.

\section{Discussion and Conclusion}
\label{section-Discussion-Conclusion}
\subsection{Challenge and Opportunities}
The past few decades have witnessed growing interests and rapid developments in  ML methods toward precision psychiatry. However, caution has been raised regarding the unrealistic hope for ML applications in clinical practice \citep{wilkinson2020time,vollmer2020machine}, and the field is still facing both conceptual and practical challenges. 
 
At the conceptual level, first, the term “disorder” was used to specifically avoid the term “disease,” which implies a level of neurobiological and physiological understanding that is lacking in psychiatry. This makes it very difficult to build clinical inference models for mental disorders since the neurological and physiological mechanisms are not always observable with sufficient precision.  As a result, it is not yet feasible to develop treatments that target underlying physiological risk factors similar to how it is possible in other areas of medicine (e.g., treating hypertension  in heart diseases).
Furthermore, each mental disorder has various types of overlapping symptoms with varying degrees, bringing an additional challenge to uniquely define the psychiatric disorder  (unlike a clear cut in cardiology or oncology).
 
Second, many disorders are presented as a spectrum, for example, autism spectrum disorder, generalized anxiety spectrum, and schizophrenia spectrum, and vary across  different patients, creating a wide range of subtypes and subject variability diagnosed with the same mental disorder. Third, due to various genetic, biochemical, neuropathological factors, the same mental disorder may have different causes and symptoms in different populations. Fourth, many mental disorders have overlapping symptoms  with other physical or mental disorders \citep{kessler2005prevalence}. For example, changes in sleep and energy level, often found in depression and generally measured using the PHQ-9 questionnaire, are very common across many other disorders. This makes it difficult to accurately diagnose mental disorders. One of the challenges of precision psychiatry is to fully dissect the mechanisms and reveal the one-to-one relationship. This can be catalyzed by rigorous and continuous measurements of factors relevant to mental health using novel data sources as described in this review.

At the practical level, many challenges remain in effective applications of ML in mental health.

\textbf{Sample size.} Datasets used in many ML applications have very small sample sizes, especially by the standard of other speech/image/video applications. Neuroimaging data collection from mental health patients is limited to one-shot examples, which brings in high signal variability in addition to the intrinsic heterogeneity or disease comorbidity. Recent developments in foundation models and their applications to mental health domain  can help overcome this challenge to some degree--for example, by sharing pretrained language models \citep{ji2021mentalbert}. However, further caution is needed to ensure appropriate validation methods on the problem-specific data. For example, studies which do not hold out all data from specific patients for validation  can lead to incorrect conclusions.
 
\textbf{Data quality.} There is a lack of standardization in data acquisition and uneven data quality, bringing up the issues of rigor and reproducibility. For example, in many studies using social media data, mental disorders self-identified by users are used as labels in the absence of clinical labels. This can lead to a post containing: “I am depressed”, being labeled as a positive class for depression regardless of the underlying clinical symptoms. Terms such as “depressed” or “anxious” have colloquial uses which can differ from clinical usage, leading to a highly noisy dataset. Furthermore, such studies also make it very difficult  to generate data for a healthy control class since a lack of mention about a disorder does not rule out its presence. Because of these reasons, the interpretation of many results reported today should be cautioned.
      
\textbf{Data privacy and security.} Advances in sensing technologies can collect a large amount of personal and sensitive data, including the location data, face images, speech conversations, and social interactions. How to collect, store and process these data without the leakage risk of privacy information remains an important challenge for advancing ML research. While research studies have oversights like internal review boards to assure the ethical use of such data, this type of data is collected in the most massive scale by large tech and social media companies. Existing regulations do not treat this data as personally identifiable information (PII) which can be used to inform the user's health. This is a major challenge in securing identifiable user data. If overcome, this will enable large-scale mental health research based on this data. To make this possible,  U.S regulations such as the Health Insurance Portability and Accountability Act (HIPAA) can be used  to govern PII acquired by all commercial entities or social media companies.
 
\textbf{Social implications and environmental factors.} Factors such as  gender and race are a critical determinant of mental health. According to WHO, mental disorders have a history of gender bias. First, in terms of gender risk factors,  females are more likely to suffer from depression and anxiety; whereas there is prevalence of autism in males. Second, in terms of gender treatment bias, doctors are more likely to diagnose depression in women compared with men, and women are more likely to be prescribed with mood altering psychotropic drugs. ML can play a role in uncovering gender or race risk factors and minimize the diagnosis or treatment bias related to these social factors.
 
\textbf{Generalizability.} The classic ML  generalization issue applies very deeply to  mental health applications, especially due to small sample sizes and poor data quality. Most studies use cross validation methods to avoid overfitting  but do not go out to collect new validation data “in the wild” to assess generalizability. Furthermore, very few studies test generalization across data sources and research institutions. For example, it is important to test how well ML models built based on speech from clinical interviews apply to non-clinical speech data.
 
\textbf{Algorithmic bias.} Digital mental health inherits the long history of bias in psychiatry, which is present at all stages of a patient journey \citep{pendse2022treatment}. This poses a major challenge in developing ML applications for mental health, making the effect of algorithmic bias possibly worse than other fields of medicine. In addition to biological underpinnings, the domains of data (such as language) also represent social underpinnings \citep{palaniyappan2021more},  and thus it is important to consider how socioeconomic factors are influencing measurements. Using training and validation sets that are representative across all demographics including race, gender, and age can not only help address some of the issues, but also uncover new  symptom expressions in various groups. This is even more important for ML approaches that inherit bias from other ML models.
 
\textbf{Interpretability.} The ability to understand which latent factors  contribute most to the outcome is the key for advancing clinical understanding of mental disorders by mental health professionals as well as for feeling the trust by the users of mental health technology. This is also an important dimension to improve “precision” in mental health. The choice of the interpretation method \citep{sheu2020illuminating}, like model-specific (such as analyzing attention weights of a transformer), or model agnostic (such as local interpretable model-agnostic explanations (LIME)), is very specific to the nature of the problem. While various interpretation methods can be used to learn about model functioning, it is important to note that the interpretation results can only be trusted as long as the challenges of generalizability and data quality are addressed. In other words, model interpretation methods can produce erratic results with insufficient or poor-quality data.  

\textbf{Causal inference.} Most ML applications in mental health  have focused on integrating more information from various data sources (as compared to a mental health professional) and reaching a diagnosis decision faster. However, diagnosing a mental disorder, even with a highly interpretable model, does not speak to the underlying causes and thus has limited implications on treating the causes. Causal inference methods supported by ML models \citep{luo2020causal} can help with “precision” treatment design \citep{prosperi2020causal}, which is the next step in the patient's journey after precision diagnosis. Recent developments in ultra-high-field neuroimaging \citep{neuner20227t} may provide a pathway for developing inference models for mental disorders by observing their underlying neural mechanisms with sufficient temporal and spatial precision.

\textbf{Clinical integration.} Many ML-based studies  have deployed limited experimental datasets. While structuring experiments and challenges for developing ML applications in mental health, it is important to consider the clinical need from a user experience perspective as well as consider key factors such as sources of data and size of held out datasets. From a user experience point of view, it is important to consider both the mental health professionals using the application \citep{rajpurkar2022ai} and their patients \citep{grande2022consumer}. Some of this work, such as running user research in various demographics, lies outside of the ML domain; however, such cross-functional research can inform best practices in developing ML models. This type of thinking with the end goal in mind is important for successful translation   of precision psychiatry research to widespread clinical practice \citep{davidson2020crossroads}. 
 
\textbf{Ethical considerations.} ML applications  in mental health raises several important ethical considerations. For example, ML models for risk assessment can lead to early screening that can in theory help with starting treatment early \citep{burr2020digital}. However,  when screening techniques are available outside clinical settings, it also creates the risk of misinterpretation by patients, which can negatively affect treatment seeking behavior or trigger self-harming thoughts in patients. Other ethical questions related to increasing the risk of self-harm arise inherently from using ML approaches that use foundation models like GPT-3 (Generative Pre-trained Transformer 3), which should be fully considered before deployment in clinical settings \citep{korngiebel2021considering}.

\subsection{Applications  of New ML Technologies}
In addition to the opportunities arising from addressing the above-mentioned challenges, precision psychiatry is also accompanied by plenty of new opportunities, especially in future ML applications.

\textbf{Data-centric approach.} In the data-driven ML view (“ML system = model/algorithm + data”), data are powerful. However, medical data are costly to collect and noisy. Currently, there is an ML paradigm shift from model-centric to data-centric, which advocates using good “small” data instead of simply collecting from big but possibly noisy data. The good quality criteria include: (i) consistency; (ii) coverage of important cases; (iii) inclusion of timely feedback from user or production data. Unlike the model-centric ML approach that focuses on modifying the model/algorithm (while fixing the data) to improve the performance, a data-centric ML approach (\url{https://datacentricai.org/}) involves building ML systems with quality data, with a goal to systematically process/augment the data (while fixing the model) to improve the ML performance \citep{polyzotis2021}. The modification of the available data may include data regeneration, data augmentation, and label refinement strategies to improve the data consistency. The process of two approaches can be iterated to bootstrap the system performance.
 
\textbf{Data augmentation approach.} To deal with the small sample size issue in the medical field, another independent ML approach is to create synthetic data (as a data augmentation strategy) for ML \citep{chen2021synthetic}. Deep learning methods such as  GAN and its variants have proved a powerful tool to generate synthetic brain images, speech, video, physiological data, and EHRs \citep{lan2020generative,geng2021deep,weldon2021generation}.
 
\textbf{Automated learning approach.} In contrast to the human-in-the-loop solutions, automated machine learning (autoML) and automated deep learning (autoDL) represent a new paradigm that aims to automate the data analysis pipeline while minimizing the need of human intervention during the course of modeling and training \citep{Chen20DL}. This has become increasingly important as the volume of social media and multimedia data streams is overwhelming and prohibitive for human effort.
 
\textbf{Data integration approach.} Integration of multimodal data is critical for psychiatric diagnostics and monitoring. Therefore, it is urgently needed to develop  weakly supervised, interpretable, multimodal deep learning pipelines to fuse histopathology, genomics, neuroimaging, and behavioral data, as well as to develop multimodal fusion algorithms for speech, video, and EHRs to assist both psychiatrists and patients. Because of the multimodality, not all data can be quantified in the Euclidean space, graph or geometric deep learning can play a role towards this research direction \citep{zhang2020deep,bronstein2021geometric}.

\subsection{Conclusion}
To date, there is still a lack of biomarkers and individualized treatment guidelines for mental illnesses. In this review, we have shown that ML technologies can be used for various stages in data analytics: detection/diagnosis, treatment selection/optimization, outcome monitoring/tracking, and relapse prevention. We predict that the multimodal integration of modernized neuroimaging, ML, genetics, behavioral neuroscience, and mobile health will open doors for new method development and technology inventions. First, making brain scans more accessible and more accurate will be the key to clinical applications of neuroimaging techniques. Using real-time fMRI, ML can guide neurofeedback-based intervention and provide closed-loop treatment or rehabilitation. As a “real-time mirror” of psychiatry, mind-control intervention can improve behavioral outcomes. Second, data-driven ML methods can be applied to identify subtypes of its symptoms and cognitive deficits, and develop model-based phenotyping \citep{habtewold2020systematic}. Third, ML methods, combined with large electronic health databases, could enable a personalized approach to psychiatry through improved diagnosis and prediction of individual responses to therapies. Fourth, when developing ML-powered technologies for psychiatry, it is important to consider concerns and feedback from  various stakeholders, including knowledgeable experts (clinical and ML experts, technology or engineer experts), decision-makers (hospital administrators, institutional leaders, state and federal government), and end users (physicians, nurses, patients, friends and family) \cite{Wiens19NatMed}. Finally, a combination of medications, wearable devices, mobile health apps, social support, and online education will be essential to improve mental health or assist therapeutic outcomes in the new era of digital psychiatry. Future precision psychiatry will leverage ML and new technologies to provide individualized custom packages that are built upon patient need and specific neural pathology.

\section*{Acknowledgments}
The research was partially supported from the US National Science Foundation (CBET-1835000 to Z.S.C.), the National Institutes of Health (R01-NS121776 and R01-MH118928 to Z.S.C.). We thank Robert MacKay for English proofreading.

\subsection*{Declaration of interests}
The authors declare no competing financial interests.

\bibliographystyle{unsrt}
\bibliography{main}

\end{document}